\documentclass[letterpaper]{article} 
\usepackage[submission]{aaai24}  
\usepackage{times}  
\usepackage{helvet}  
\usepackage{courier}  
\usepackage[table]{xcolor} 

\usepackage[hyphens]{url}  
\usepackage{graphicx} 
\usepackage{natbib}  
\usepackage{caption} 
\usepackage{algorithm}
\usepackage{algorithmic}
\usepackage{graphicx}
\usepackage{bm}

\usepackage{amsthm}
\usepackage{amsmath}
\usepackage{amsfonts}
\theoremstyle{definition}
\newtheorem{definition}{Definition}

\newtheorem*{problem*}{Problem}
\newtheorem{theorem}{Theorem}

\usepackage{booktabs}
\usepackage{multirow}
\usepackage{xspace}
\usepackage{xcolor}
\usepackage{newfloat}
\usepackage{listings}
\usepackage{subfigure}
\DeclareCaptionStyle{ruled}{labelfont=normalfont,labelsep=colon,strut=off} 
\floatstyle{ruled}
\newfloat{listing}{tb}{lst}{}
\floatname{listing}{Listing}

\newcommand{\statespace}{\ensuremath{\mathbf X}\xspace}
\newcommand{\obsspace}{\ensuremath{\mathbf Z}\xspace}
\newcommand{\actspace}{\ensuremath{\mathbf U}\xspace}
\newcommand{\imgcontroller}{\ensuremath{\emph{h}}\xspace}
\newcommand{\dynmodel}{\ensuremath{f}\xspace}
\newcommand{\obsmodel}{\ensuremath{o}\xspace}
\newcommand{\initstate}{\ensuremath{x_0}\xspace}
\newcommand{\safeprop}{\ensuremath{\varphi}\xspace}

\newcommand{\obsseqspace}{\ensuremath{\obsspace^m}\xspace}
\newcommand{\dataset}{\ensuremath{\mathbf{Z}}\xspace}
\newcommand{\dataseq}
{\ensuremath{\mathbf{z}}\xspace}
\newcommand{\labelpred}
{\ensuremath{\rho}\xspace}
\newcommand{\chancepred}
{\ensuremath{g}\xspace}

\newcommand{\evaluator}
{\ensuremath{v}\xspace}
\newcommand{\imgforecaster}
{\ensuremath{f_{g}}\xspace}
\newcommand{\latforecaster}
{\ensuremath{f_{l}}\xspace}
\newcommand{\enc}
{\ensuremath{e}\xspace}
\newcommand{\dec}
{\ensuremath{d}\xspace}
\newcommand{\latspace}
{\ensuremath{\Theta}\xspace}
\newcommand{\latvec}
{\ensuremath{\theta}\xspace}

\title{How Safe Am I Given What I See?\\ Calibrated Prediction of Safety Chances for Image-Controlled Autonomy }

\author {
    Zhenjiang Mao,
    Carson Sobolewski,
    Ivan Ruchkin
}
\affiliations {
    University of Florida\\
    z.mao@ufl.edu, csobolewski@ufl.edu, iruchkin@ece.ufl.edu
}
\usepackage{bibentry}

\begin{document}

\maketitle

\begin{abstract}%
End-to-end learning has emerged as a major paradigm for developing autonomous controllers. Unfortunately, with its performance and convenience comes an even greater challenge of safety assurance. A key factor in this challenge is the absence of low-dimensional and interpretable dynamical states, around which traditional assurance methods revolve. Focusing on the online safety prediction problem, this paper systematically investigates a flexible family of learning pipelines based on generative world models, which do not require low-dimensional states. To implement these pipelines, we overcome the challenges of missing safety labels under prediction-induced distribution shift and learning safety-informed latent representations. Moreover, we provide statistical calibration guarantees for our safety chance predictions based on conformal inference. An extensive evaluation of our predictor family on two image-controlled case studies, a racing car and a cartpole, delivers counterintuitive results and highlights open problems in deep safety prediction. \end{abstract}

\section{Introduction}

\looseness=-1
To handle the complexity of the real world, autonomous systems increasingly rely on high-resolution sensors such as cameras and lidars, as well as large learning architectures to process the sensing data. It has become increasingly commonplace to use end-to-end reinforcement and imitation learning~\cite{codevilla_end--end_2018,tomar_learning_2022,betz_autonomous_2022}, conveniently bypassing conventional intermediate components such as state estimation and planning. While end-to-end learning can produce sophisticated behaviors from large raw datasets, it complicates safety assurance of such autonomous systems~\cite{fulton_formal_2019}.  

Traditional approaches to ensure safety (e.g., that a car does not hit an obstacle) predominantly rely on the notion of a \emph{low-dimensional state} with a physical meaning (e.g., the car's position and velocity). For example, reachability analysis propagates the system's state forward
~\cite{bansal_hamilton-jacobi_2017,chen_reachability_2022}, barrier functions synthesize safe low-dimensional controllers~\cite{ames_control_2019,xiao_safe_2023}, and trajectory predictors output future states of agents~\cite{salzmann_trajectron_2020,teeti_vision-based_2022}. These methods do not straightforwardly scale to image-based controllers with thousands of inputs without an exact physical interpretation. Instead, model-based approaches require abstractions of sensing and perception to take images into account. These abstractions are either system-specific and effortful to build (e.g., modeling the ray geometry behind pixel values~\cite{santa_cruz_nnlander-verif_2022}) or simplified and potentially inaccurate (e.g., a linear overapproximation of visual perception~\cite{hsieh_verifying_2022}).

The problem raised in this paper is the reliable online prediction of safety for an autonomous system with an image-based controller \emph{without access to a physically meaningful low-dimensional dynamical state or model}. For instance, given an image, what is the probability that a racing car stays within the track's bounds within 5 seconds after the image was taken? In this context, the reliability requirement, also known as \emph{calibration}~\cite{guo_calibration_2017}, would be to provide an upper bound on the difference between the estimated and true probability. Devoid of typical dynamical models, this setting calls for a careful combination of image-based prediction and statistical guarantees. Existing works focus on learning to control rather than assuring image-based systems~\cite{hafner_dream_2019}, assume access to a true low-dimensional initial state~\cite{lindemann_conformal_2023}, or do not provide any reliability guarantees~\cite{acharya_competency_2022}. 
Nonetheless, reliably solving the  prediction problem would enable downstream safety interventions like handing the control off to a human or switching to a fallback controller, which are outside of this paper's scope. 

\looseness=-1
This paper proposes a \emph{family of learning pipelines} for the formalized safety prediction problem, shown in Fig.~\ref{fig:overview}. The proposed family offers flexible modularity (e.g., whether it uses intermediate representations) and tunable specificity to a particular controller. Some of these pipelines repurpose the latest generative architectures for reinforcement learning, known as \emph{world models}~\cite{ha_recurrent_2018}, for online prediction of future images from the recent ones.  
World models tend to produce distribution-shifted image forecasts, which poses a challenge to evaluating the safety of these forecasts. We address this challenge by using robust vision features and data augmentation --- and also investigate using safety labels to regularize latent representations learned by the world models. 

To provide calibration guarantees, we combine post-hoc calibration~\cite{zhang_mix-n-match_2020} with the recently popularized distribution-free technique of \emph{conformal prediction}~\cite{vovk_algorithmic_2005,lei_distribution-free_2018}. This enables us to tune the predictive safety probabilities orthogonally to the design choices within the prediction pipeline and provide statistical bounds for them from validation data. 

We perform extensive experiments on two popular benchmarks in the OpenAI Gym~\cite{brockman_openai_2016}: racing car and cart pole. We particularly focus on long-horizon predictions, which leads us to counterintuitive results on the impact of modularity and controller-specificity  in our family of learning-based prediction pipelines. Our experiments also show that predicting well-calibrated safety chances is easier than safety labels over longer horizons. 

In summary, this paper makes three contributions: 
\begin{enumerate}
    \item A flexible family of learning pipelines for online safety prediction in image-controlled autonomous systems. 
    \item A conformal post-hoc calibration technique with statistical guarantees on safety chance predictions. 
    \item An extensive experimental evaluation of our predictor family on two case studies. 
\end{enumerate}

After describing the problem and notation in Sec.~\ref{sec:prelim}, we introduce our family of prediction approaches in Sec.~\ref{sec:approach}. The case studies and their results are presented in Sec.~\ref{sec:results}, after which we review the related work in Sec.~\ref{sec:relwork} and conclude. 

\section{Preliminaries}\label{sec:prelim}

This section introduces the necessary notation and describes the problem addressed in this paper.

\subsection{Problem Setting}

\begin{definition}[Dynamical system]
A discrete-time \emph{dynamical system} $s=(\statespace,\obsspace,\actspace,\imgcontroller,\dynmodel,\obsmodel,\initstate,\safeprop)$ consists of:
\begin{itemize}
	\item \emph{State space} \statespace, containing continuous states $x$ 
	\item \emph{Observation space} \obsspace, containing images $y$
	\item \emph{Action space} \actspace, with  discrete/continuous commands $u$
 	\item \emph{Image-based controller} $\imgcontroller:\mathbf \obsspace \rightarrow \mathbf \actspace$, typically implemented by a neural network
     \item \emph{Dynamical model} $\dynmodel: \statespace \times \actspace \rightarrow \statespace$ , which sets the next state from a past state and an action (unknown to us)
      \item \emph{Observation model} $\obsmodel:\statespace \rightarrow \obsspace$, which generates an observation based on the state  (unknown to us)
      \item \emph{Initial state} $ \initstate$, from which the system starts executing
       	\item \emph{State-based safety property} $\safeprop: \mathbf{X} \rightarrow \{0,1\}$, which determines whether a given state $x$ is safe
        \end{itemize}
\end{definition}

We focus on systems with observation spaces with thousands of pixels and unknown non-linear dynamical and observation models. In such systems, while the state space \statespace is conceptually known (if only to define \safeprop), it is not necessary (and often difficult) to construct \dynmodel and \obsmodel because the controller acts directly on the observation space \obsspace. Without relying on \dynmodel and \obsmodel, end-to-end methods like deep reinforcement learning \cite{mnih2015human} and imitation learning~\cite{hussein_imitation_2017} are used to train a controller \imgcontroller by using data from the observation space. Once the controller is deployed in state \initstate, the system executes a  \emph{trajectory}, which is a sequence $\{x_i,y_i,u_i\}^{t}_{i=0}$  up to time $t$, where: 
\begin{align}
x_{i+1} = \dynmodel(x_i, u_i), \quad  y_i = \obsmodel(x_i), \quad 
u_i = \imgcontroller(y_i)
\end{align}


Instead of using this model, we extract predictive information from three sources of data. First, we will use the current observation, $y_i$, at some time $i$. 
For example, when a car is at the edge of the track, it has a higher probability of being unsafe in the next few steps. 
Second, past observations $y_{i-m+1},\dots, y_i$ provide dynamically useful features that can only be extracted from time series, such as the speed and direction of motion. For example, just before a car enters a turn, the sequence of past observations implicitly informs how hard it will be to remain safe. 
Third, not only do observations inform safety, but so do the controller's past/present outputs $\imgcontroller(y_{i-m+1},\dots, y_i)$. For instance, if by mid-turn the controller has not changed the steering angle, it may be less likely to navigate this turn safely.  

Our goal is to predict the system's safety $\safeprop (x_{i+k})$ at time $i+k$ 
given a series of $m$ observations $\mathbf y_i=(y_{i-m+1},...,y_i)$. 
This sequence of observations does not, generally, determine the true state $x_i$ (e.g., when $m=1$, function \obsmodel may not be invertible). This leads to \emph{partial state observability}, which we model stochastically. Specifically, we say that from the predictor's perspective, $x_i$ is drawn from some belief distribution $\mathcal{D}_{\mathbf y_i}$. This induces a distribution of subsequent trajectories and transforms future safety $\safeprop (x_{i+k})$ into a Bernoulli random variable. Therefore, we will estimate the conditional probability $P(\safeprop (x_{i+k}) \mid \mathbf y_i)$ and provide an error bound on our estimates. Note that process noise in $f_d$ and measurement noise in $f_o$ are orthogonal to the issue of partial observability; nonetheless, both of these noises are supported by our approach and would be treated as part of the stochastic uncertainty in the future $\safeprop$. To sum up the above, we arrive at the following problem description. 

\begin{problem*}[Calibrated safety prediction] Given horizon $k>0$, confidence $\alpha \in (0, 0.5)$, and observations $\mathbf y_i$ from some system $s$ with unknown \dynmodel and \obsmodel, estimate future safety chance $P(\safeprop(x_{i+k})\mid \mathbf y_i)$ and provide an upper bound for the estimation error that holds in at least $1-\alpha$ cases. 
\end{problem*}


\begin{figure}
	\centering         \includegraphics[width=\columnwidth]{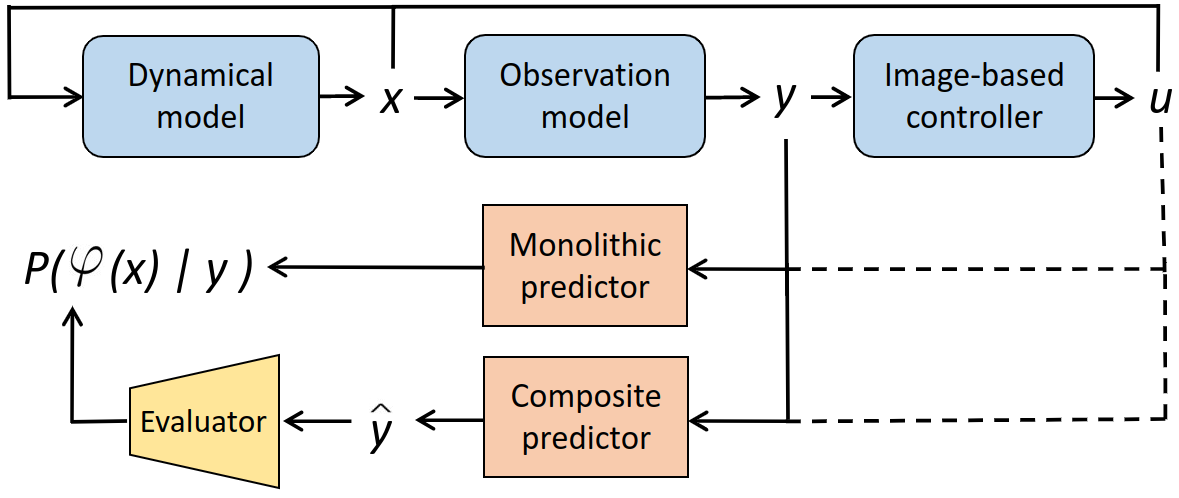}
	\caption{Dynamical system with predictors. Arrows show data flow, and dashes show optional controller dependence.}
	\label{fig:overview}
 \end{figure}


\subsection{Predictors and Datasets}

To address the above problem we will build two types of predictors: for safety labels and for safety chances. 

\begin{definition}[Safety label predictor]  \label{def:predictor-label}
  For horizon $k>0$, a \emph{safety label predictor} $\labelpred: \obsseqspace \rightarrow \{0, 1\} $ predicts the safety  $\safeprop (x_{i+k})$ at  time $i+k$.
\end{definition}

\begin{definition}[Safety chance predictor] 
\label{def:predictor-chance}
  For horizon $k>0$, a \emph{safety chance predictor} $\chancepred: \obsseqspace \rightarrow [0, 1] $ predicts the safety chance $P(\safeprop (x_{i+k})\mid \mathbf{y}_i )$ at  time $i+k$.
\end{definition}
  


\emph{Controller-specific} (\emph{controller-independent}) predictors are trained on observation-controller (observation-action) datasets respectively. Using controller-independent predictors trades off some prediction power for generalizability --- and it is the first flexibility dimension (i.e., \textbf{controller-specificity}) in our family of learning pipelines. We compare these two types of predictors in Sec.~\ref{sec:results}.


\looseness=-1
Our problem setting assumes that all our training data for training these predictors is obtained offline. To train these predictors, we collect two types of datasets, with a conveniently unified notation \dataset and \dataseq to mean either dataset and its elements. 
The first type in Def.~\ref{def:ocd} only contains data from a specific controller, while the second in Def.~\ref{def:oad} can mix controllers because due to explicit actions. 

\begin{definition}[Observation-controller dataset]\label{def:ocd}
  An \emph{observation-controller dataset} $\dataset=\{ (\dataseq_j,\safeprop_j) ~|~ j =1 ,\dots, N\}$ consists of pairs of $m$-long sequences $\dataseq_j:=(y_{i-m},...,y_i) $ for some time $i$ and safety labels $\safeprop_j := \safeprop(x_{i+k})$ at $k$ steps later, collected by executing a fixed controller \imgcontroller.
\end{definition}  

\begin{definition}[Observation-action dataset]\label{def:oad}
  An \emph{observation-action dataset} $\dataset=\{ (\dataseq_j,\safeprop_j) ~|~ j =1 ,\dots,N\}$  consists of pairs of $m$-long sequences of paired observations and corresponding actions $\dataseq_j=((y_{i-m+1}, u_{i-m+1}),...,(y_i, u_i))$, and safety labels $\safeprop_j := \safeprop(x_{i+k})$ obtained $k$ steps later.
\end{definition}










\section{Safety Predictor Family}\label{sec:approach}













First, we describe how we develop safety \emph{label} predictors. Then we transform them into safety \emph{chance} predictors. 

\subsection{Monolithic and Composite Label Predictors}

\begin{figure}[t]
    \centering
    \includegraphics[width=\columnwidth]
    {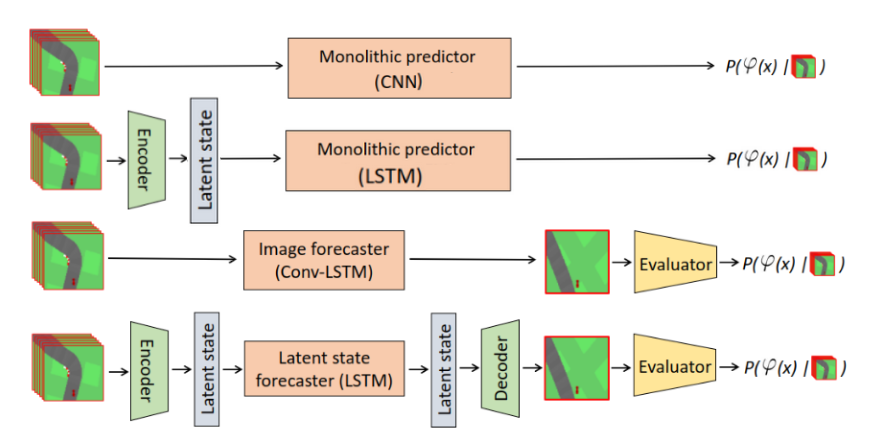}
    \caption{Our predictors (from top to bottom):  monolithic, monolithic latent, composite image-based, and composite latent-based. 
    }
    \label{fig:predictors}
\end{figure}

The second dimension of our family of learning pipelines is \textbf{modularity}. In this dimension, we  distinguish three types of safety label predictors: monolithic predictors, composite image predictors, and composite latent predictors. The high-level distinction between the monolithic and composite predictors is that the latter end with an \emph{evaluator}, a separate component that determines the safety of the prediction --- as shown in Fig.~\ref{fig:overview}.  The predictor structures are summarized in Fig.~\ref{fig:predictors} and explained below.


Monolithic predictors directly determine the future safety property based on the observations, as described in Def.~\ref{def:predictor-label}. 
These predictors are essentially binary classifiers that leverage deep vision models (with convolutional layers) and time-series models (with recurrent layers). These models are trained in a supervised way on any training dataset $\dataset_t$ with a given horizon $k$ with a typical binary classification loss, such as cross-entropy. 
The key drawback of monolithic predictors is their need to be completely re-trained to change the prediction horizon or tune the hyperparameters.

Typically, a Convolutional Neural Network (CNN) serves as an effective classifier for safety prediction. However, as the prediction horizon increases, CNNs may prove inadequate for processing sequential data. Recurrent Neural Networks, such as Long Short-Term Memory (LSTM) networks ~\cite{lindemann_survey_2021}, are burdened by computational demands in such scenarios. To address this, we have incorporated a \textit{Variational Autoencoder} (VAE)~\cite{kingma_auto-encoding_2014} into the LSTM. The VAE compresses the image sequence into a latent sequence, and an LSTM is employed as the safety predictor. This hybrid model aims to optimize both the representation learning capabilities of VAE and the sequential processing prowess of LSTM for enhanced predictive performance. This kind of monolithic latent predictor is defined as below:
\begin{definition}[Monolithic latent predictor]
A \emph{monolithic latent predictor} consists of two parts: an autoencoder and an LSTM. The \emph{autoencoder} provides an encoder \enc and decoder \dec components that define a latent space \latspace containing state vectors \latvec. 
The \emph{LSTM} works as the safety label predictor (Def.~\ref{def:predictor-label}) from the latent space \latspace. 
\end{definition}

A composite predictor consists of multiple learning models. A \emph{forecaster} model constructs the likely future observations, essentially approximating the dynamics \dynmodel and observation model \obsmodel. Another binary classifier --- an \emph{evaluator} --- judges whether a forecasted observation is safe. 

\begin{definition}[Composite image predictor]
A \emph{composite image predictor} consists of two parts: an image forecaster and an evaluator. The \emph{image forecaster} $\imgforecaster: \obsseqspace \rightarrow \mathbf{Y}$ predicts the future observation $\hat{y}_{i+k}$ based on past $\mathbf{z}_i$, 
and the \emph{evaluator} $e: \mathbf{Y} \rightarrow \{0, 1\}$ determines the safety of the predicted images 
$\varphi (\hat{y}_{i+k})$. The composite prediction is  $\evaluator(\imgforecaster(\mathbf{z}_i))$. 
\end{definition}
Our implementation of the composite image predictor will use a \emph{convolutional LSTM} (conv-LSTM)~\cite{shi_convolutional_2015} as an image forecaster, and a \emph{convolutional neural network} (CNN) for the evaluator. 

\begin{definition}[Composite latent predictor]
A \emph{composite latent predictor} consists of three parts: an autoencoder, a latent forecaster, and an evaluator. The \emph{autoencoder} provides an encoder \enc and decoder \dec components that define a latent space \latspace containing state vectors \latvec. 
The \emph{latent forecaster} $\latforecaster: \latspace^m \rightarrow \latspace$ predicts the future latent state $\hat{\latvec}_{i+k}$ based on past $m$ latent states. 

Same as for the image predictor, the \emph{evaluator} $\evaluator: \mathbf{Y} \rightarrow \{0, 1\}$ operates over the images obtained from the predicted latent states by the decoder. The composite prediction is  $\evaluator(\dec(\latforecaster(\enc(\mathbf{z}_i))))$. 
\end{definition}

We implement the encoder/decoder using a \emph{Variational Autoencoder} (VAE)~\cite{kingma_auto-encoding_2014}. The latent forecasting  is performed using a \emph{Long-Short Term Memory} (LSTM) network~\cite{lindemann_survey_2021}, and the evaluator, as above, is a CNN. 



\subsection{Training Process}

 
Monolithic predictors \labelpred, evaluators \evaluator, and image forecasters \imgforecaster are trained in a supervised manner on a training dataset $\dataset_{t}$ containing observation sequences with associated safety labels or future images respectively. Latent forecasters \latforecaster are trained on future latent vectors, obtained from true future observations with a VAE encoder \enc, which is trained as described below. 
The mean squared error (MSE) loss is implemented for the LSTM latent forecaster and conv-LSTM image forecaster, while the monolithic predictors and evaluators use cross entropy (CE) as the loss function. 
We also implement the early stopping strategy to reduce the learning rate as the total loss drop slows --- and stop the training when it reaches below a given threshold. 

For the VAE, the total loss $\mathcal{L}$ consists of three parts. The first is the reconstruction loss $\mathcal{L}_{recon}$, which quantifies how well an original image $y$ is approximated by $\dec(\enc(y))$ using the MSE loss. The second is the latent loss $\mathcal{L}_{latent}$, which uses 
Kullback-Leibler divergence loss 
to minimize the difference between two latent-vector probability distributions 
$\enc(y\mid \theta)\approx \dec(\theta \mid y)$, where $\theta$ is a latent vector and $y$ is an input. 
In order to preserve the information about safety in latent state representations, we add an optional safety loss $\mathcal{L}_{safety}$, which is a cross entropy loss based on the true safety $\safeprop_i$ truth of the original images $y_i$ and the safety evaluation of the reconstructed images: 
\begin{equation}\label{eq:vae-loss}
\mathcal{L}_{safety}=\mathcal{L}_{CE}(\evaluator(\dec(\enc(y_i))),\safeprop_i)
\end{equation}

So the total loss function for VAE combines the three parts with regularization parameters $\lambda_1$, $\lambda_2 > 0$:

\begin{equation}\label{eq:safety-loss}
\mathcal{L}=\mathcal{L}_{recon}+\lambda_1\mathcal{L}_{latent}+\lambda_2\mathcal{L}_{safety}
\end{equation}

One challenge is that the safety label balance changes in $\dataset_{t}$ with the prediction horizon $k$, leading to imbalanced data for higher horizons. To ensure a balanced label distribution, we resample with replacement for a 1:1 safe:unsafe class balance both in the training and testing datasets.  

\begin{figure}[t]
	\centering
 \includegraphics[width=\columnwidth]{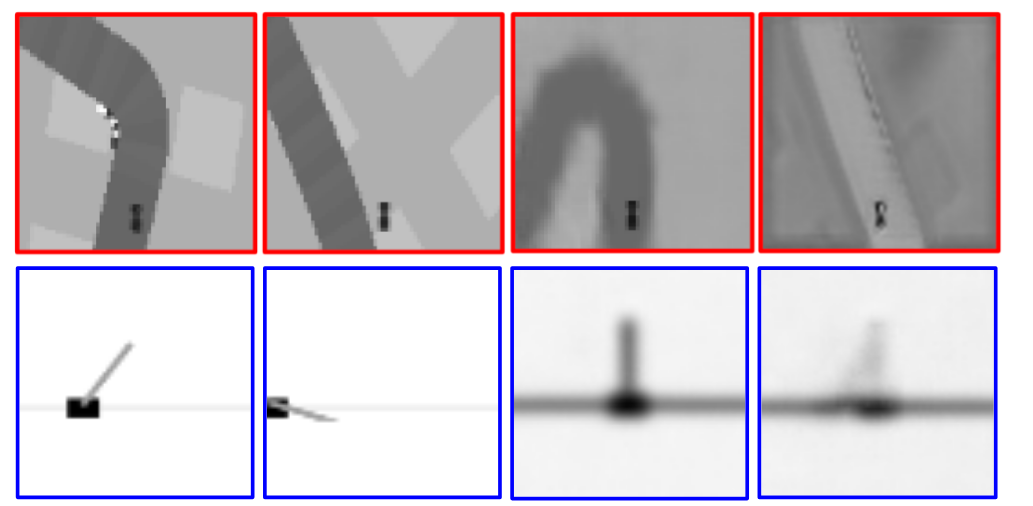}

	\caption{Upper: Car observations, L to R: safe car from \obsspace, unsafe car from \obsspace, safe car from \dec(\latforecaster(\enc(\obsspace))), safe car from \imgforecaster(\obsspace).  Lower: Cart pole observations. In both rows, the two rightmost images are distribution-shifted.} 
    	\label{fig:car-obs}
\end{figure}

Another challenge is the \emph{distribution shift} between the original and forecasted images produced by label predictor \labelpred. The issue is that the forecasted images are distorted (e.g., see the two rightmost images in Fig.~\ref{fig:car-obs}) and they do not automatically come with safety labels because they are sampled from the forecasted space without a physical ground truth. However, we need to train high-performance evaluators on the forecasted images, as per Fig.~\ref{fig:predictors}. 

\looseness=-1
To overcome the distribution shift, we implement two specialized evaluators to finish the last step that generates binary safety consequences from predicted images. One uses a vision approach based on robust domain-specific features of the image. For example, for the racing car case study, we take advantage of the fixed location of the car in the image and the contrasting colors to determine whether the car is in a safe position. 
Specifically, we crop the image to the area directly surrounding the car and use the mean of the pixel values to determine whether the car is on or off the track. To resolve issues with this approach on inverted-color images, such as the rightmost picture in Fig.~\ref{fig:car-obs}, we use the median pixel value of the whole image to determine whether the track surface is painted in an inverted, lighter-than-background color and then determine the safety accordingly. The other evaluator is trained with data augmentation. Our training process randomly adjusts the brightness, inverts the input image, and adds a Gaussian blur to it.


\subsection{Conformal Calibration for Chance Predictors}


To turn a label predictor \labelpred into a chance predictor \chancepred, we take its normalized softmax outputs and perform post-hoc calibration~\cite{guo_calibration_2017,zhang_mix-n-match_2020}. On a  held-out calibration dataset $\dataset_c$, we hyperparameter-tune over state-of-the-art post-hoc calibration techniques: temperature scaling, logistic calibration, beta calibration, histogram binning, isotonic regression, ensemble of near isotonic regression (ENIR), and Bayesian binning into quantiles (BBQ). By selecting the model that has minimum ECE, we can get calibrated softmax values $\chancepred(y_i)$. Furthermore, to ensure that samples are spread evenly across bins to support our statistical guarantees, we perform \emph{adaptive binning} as defined below to construct a validation dataset $\dataset_v$, on which we obtain our guarantees.


\begin{definition}[Adaptive binning]\label{def:binning}
Given a dataset \dataset, split $\dataset$ into $Q$ bins $\{B_{j}\}_{j=1}^{Q}$ by a constant count of samples, $ \lfloor |\dataset|/Q \rfloor $ each. The resulting $\{B_{j}\}_{j=1}^{Q}$ 
is a \emph{binned dataset}.
\end{definition}

From each bin $B_j$
, we draw $N$ i.i.d. samples with replacement  $M$ times to get $\{B_{j1},\ldots,B_{jM}\}$. 
For each resampled bin $B_{ji}$, we calculate the average safety confidence score
$\overline{\chancepred}_{ji}$, the true safety chance $p_{ji}$ (i.e., the fraction of predictions that are indeed safe), and the calibration error 
$\delta_{ji} := 
| \overline{\chancepred}_{ji} - p_{ji} |$. 
Given a bin number $j$, our goal is to build \textit{prediction intervals}  $[0, c_j]$ that contain the calibration error 
of  the next (unknown) average confidence $\overline{\chancepred}_{j*}$ that falls into bin $B_j$ with chance at least $1- \alpha$:
\begin{equation}
P(|\chancepred_{j*} - p_j | \leq c_j)\geq 1-\alpha \text{ for } j=1\ldots Q,
\label{eq:coverage-guarantee}
\end{equation}
where $\alpha$ is the miscoverage level.

\begin{algorithm}[t]
\caption{Conformal calibration for chance predictions}
\label{alg:algorithm}

\textbf{Input}: A validation dataset bin $B=\{b_k\}_{k=1,...}$ which each contains sequences of observations and safety $b_k=({\mathbf{y_k}},\mathbf{\varphi ( x_k) })$ , trained safety chance predictor \chancepred and miscoverage level $\alpha$ \\

\textbf{Output}: confidence bound $c$ satisfying Eq.~\ref{eq:coverage-guarantee}. 

\textsc{Function} \texttt{ConCali} ($B,g,\alpha$):

\begin{algorithmic}[1] 

\FOR{$j=1$ to $Q$}

\FOR{$i=1$ to $M$}

\STATE  $B_{i} \gets N$ i.i.d. samples from $B$  \COMMENT{resampled bin}

\STATE 
$\overline{q}_{i} \gets \frac{1}{N}\sum^{N}_{l=1}\chancepred(y_{l}), \text{ for each } y_{l} \in B_{i} $ 

\COMMENT{mean safety chance prediction}

\STATE $\overline{p_{i}}  \gets  \frac{1}{N}\sum^{N}_{l=1} \safeprop(x_l), \text{ for each } \safeprop(x_l) \in B_{i} $ \COMMENT{true safety chance}




\STATE $\delta_{i} \gets |\overline{q}_{i}- \overline{p_{i}}|$ \COMMENT{non-conformity score}

\ENDFOR

\STATE $n \gets \lceil (M+1)(1-\alpha) \rceil $ \COMMENT{conformal quantile} 

\STATE $c \gets $ the $n$-th smallest value among $\delta_{1}, \ldots, \delta_{M}$ 

\ENDFOR

\STATE \textbf{return} $c$

\end{algorithmic}
\end{algorithm}

That is, we aim to predict a statistical upper bound $c_j$ on the error of our chance predictor in each bin. 
After we obtain a chance prediction $g_{j*}$, it will be turned into an uncertainty-aware interval $[g_{j*} - c_j; g_{j*} + c_j]$, which contains the true probability of safety in $1-\alpha$ cases. Notice that this guarantee is relative to the binned dataset fixed in Def.~\ref{def:binning}. 

To provide this guarantee, we apply
\textit{conformal prediction} ~\cite{lei_distribution-free_2018}  in Alg.~\ref{alg:algorithm}. Intuitively, we rank the safety chance errors in resampled bins and obtain a statistical upper bound on this error for each bin. Similarly to existing works relying on conformal prediction~\cite{qin_statistical_2021,lindemann_conformal_2023}, this algorithm guarantees Eq.~\ref{eq:coverage-guarantee}. Note that bin averaging is necessary to obtain meaningful probabilities, rather than binary outcomes.  

\begin{figure*}[th!]
\centering
\includegraphics[width=1\textwidth]{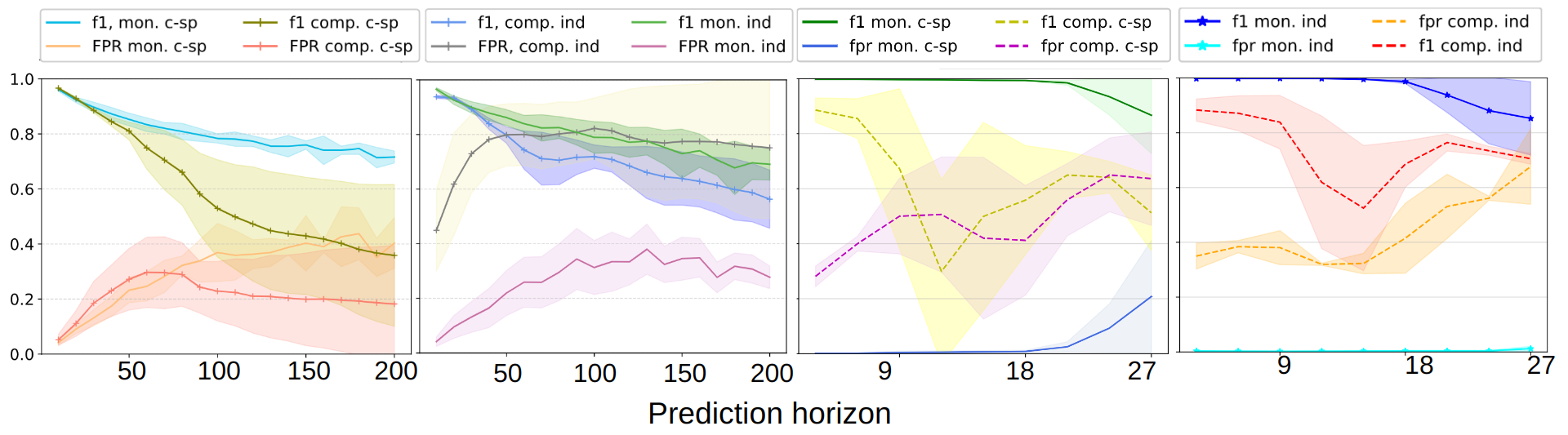} 
\caption{Performance of safety label predictors over varied horizons. L to R: (1) controller-specific (`c-sp') monolithic (`mon') vs. latent composite (`comp') for the racing car; (2)  controller-independent (`ind') monolithic vs. latent  composite for the racing car; (3) controller-specific monolithic vs. latent  composite for the cart pole;  (4) controller-independent  monolithic vs. latent composite for the cart pole. Shaded uncertainty shows standard deviation due to different controllers and resampling.}
\label{fig:results-label-predictors}
\end{figure*}

\begin{theorem}
Theorem 2.1 in (Lei et al.
2018). Given a  dataset bin $B= \{b_{k}\}_{k=1}^K$ of i.i.d. observation-state pairs $b_{k}=(y_k,x_k)$, 
we obtain a collection of datasets $\{B_{j}\}^{M}_{j=1}$ by drawing $M$ datasets of $N$ i.i.d. samples from $B$, leading to datasets $B_{j}$ to be drawn i.i.d. from a dataset distribution $\mathcal{D}$. Then for another, unseen dataset $B_{M+1} \sim \mathcal{D}$, safety chance predictor $g$, and miscoverage level $\alpha$, calculating $c = \texttt{ConCali} (B,g,\alpha)$ leads to prediction intervals with guaranteed containment:  $$P_\mathcal{D}(|\overline{q}(B_{M+1})- \overline{p}(B_{M+1})| \leq c)\geq 1-\alpha, $$ 
where $\overline{q}(B_{M+1})$ is the mean safety chance prediction in $B_{M+1}$: $$\overline{q}(B_{M+1}) \gets \frac{1}{N}\sum^{N}_{l=1}\chancepred(y_{l}), \text{ for each } y_{l} \in B_{M+1} $$ and $\overline{p}(B_{M+1})$ is the mean true safety chance in $B_{M+1}$: $$\overline{p}(B_{M+1})  \gets  \frac{1}{N}\sum^{N}_{l=1} \safeprop(x_l), \text{ for each } \safeprop(x_l) \in B_{M+1} $$ where $\safeprop(x)$ is the safety of $x$.

\end{theorem}
\section{Experimental Results}\label{sec:results}




\paragraph{Systems}
The racing car and cart pole from the OpenAI Gym~\cite{brockman_openai_2016} are selected as our study cases. These environments are more suitable than pre-collected autonomy datasets, such as the Waymo Open Dataset, for two reasons. First, to study safety prediction, we require a large dataset of safety violations, which is rarely found in real-world data. Second, the simulated dynamical systems are convenient to collect an unbounded amount of data with direct access to the images and the ground truth for evaluation (e.g., the car's position).

We defined the racing car's safety as being located within the track surface. The cart pole's safety is defined by the angles in the safe range of $[-6, 6]$ degrees, whereas the whole activity range of the cart pole is $[-48, 48]$ degrees.

\paragraph{Performance Metrics}
We use the \emph{F1 score} as our main metric for evaluating safety label predictors, as it balances the precision and recall: $F1 = 2 \times \frac{\text{Precision} \times \text{Recall}}{\text{Precision} + \text{Recall}} = 2 \times \frac{TP}{{TP + \frac{FP + FN}{2}}}$. 
False positives are also a major concern in safety prediction (actually unsafe situations predicted as safe), so we also evaluate the \emph{False Positive Rate} (FPR):  $\text{FPR} = \frac{FP}{FP + TN}$. 
To evaluate our chance predictors, we compute the \emph{Expected Calibration Error} (ECE, intuitively a weighted average difference between the predicted and true probabilities) and \emph{Maximum Calibration Error} (MCE, intuitively the maximum aforementioned difference)~\cite{guo_calibration_2017,minderer_revisiting_2021}. 

\subsection{Experimental Setup}

\subsubsection{Hardware}
\looseness=-1
The computationally heavy training was performed on a single NVIDIA A100 GPU, and other light tasks like data collection and conformal calibration were done on a CPU workstation with 12th Gen Intel Core i9-12900H CPU and 64GB RAM in Ubuntu 22.04.

\subsubsection{Dataset}

Deep Q-networks (DQN) were used to implement image-based controllers both for the racing car and cart pole. For racing car, we collected 240K samples for training and 
60K for calibration/validation/testing each.  
For the cart pole, we collected 90K samples for each of the four datasets. 
All images were processed with normalization and grayscale conversion.

Each case study's data is randomly partitioned into four datasets. 
All the training tasks for the predictors are performed on the training dataset $\dataset_t$. The calibration dataset $\dataset_c$ is to tune predictor hyperparameters and fit the  calibrators. The validation dataset $\dataset_v$ is used to produce conformal guarantees, and finally the test dataset $\dataset_e$ is used to compute the performance metrics like F1, FPR, ECE and brier scores. 


\subsubsection{Training details}

\looseness=-1
Pytorch 1.13.1 with the Adam optimizer was used for training. The maximum training epoch is 500 for VAEs and 100 for predictors.
The safety loss in Eq.~\ref{eq:safety-loss} uses $\lambda_1 =1$ and $ \lambda_2 = 4096$, which equals the total pixel count in our images. The miscoverage level is $\alpha=0.05$. The remaining hyperparameters are found in the appendix. We released our code at \url{https://github.com/maozj6/hsai-predictor}.



\subsection{Comparative Results}

Below we present comparisons and ablations along the four key dimensions of the proposed family of predictors and also discuss our conformal calibration's performance. We use the following abbreviations in the figures and tables: `mon',`comp',`c-sp' and `ind' stand for monolithic, composite, controller-specific, and controller-independent respectively. 

\paragraph{Result 1: Monolithic predictors outperforms composite ones.}

\looseness=-1

Monolithic predictors do not learn the underlying dynamics, so we hypothesized that they would do well on short horizons --- but lose to composite predictors on longer horizons. 
To our surprise, as illustrated in the Fig.~\ref{fig:results-label-predictors}, the performance of composite predictors degrades faster for longer horizons; the only aspect in which composite predictors excelled was a better FPR for the racing car, which may be desirable in safety-critical systems. We attribute the degradation of composite predictors to the challenge of learning coherent long-term latent dynamics, which remains an open research problem.

\paragraph{Result 2: Latent predictors exceed the performance of image-based ones.}

\looseness=-1
Latent predictors outperform image-based ones both in F1 and FPR (as shown in the Tabs.~\ref{tab:f1car},~\ref{tab:fprcar},~\ref{tab:f1pole}, and~\ref{tab:fprpole} in the appendix), for two reasons. The first is that the efficient compression by a safety-informed VAE supports generalizable learning of the dynamics: note the performance drop of image-based predictors when the horizon exceeds the training sequence length. Second,  
image-based forecasters tend to induce a stronger distribution shift on the forecasted images, hence disrupting the evaluator.



\paragraph{Result 3: Controller-independent and controller-specific predictors show comparable performance.}

Our initial hypothesis was that controller-specific predictors would work better due to less variance. For composite predictors, the results are inconsistent with our hypothesis --- see 1st vs 2nd, and 3rd vs 4th plots in Fig.~\ref{fig:results-label-predictors} (and also Tabs.~\ref{tab:f1car},~\ref{tab:fprcar},~\ref{tab:f1pole}, and~\ref{tab:fprpole} in the appendix). 
For monolithic predictors, we were surprised to find no significant difference between controller-specific and controller-independent ones in F1 scores, while the independent ones do slightly better in FPR. 
This means that monolithic predictors obtain their versatility virtually ``for free''.


\paragraph{Result 4: Calibrated predictors are superior to uncalibrated ones.} 

\begin{figure}[t]
\centering
\includegraphics[width=1\columnwidth]{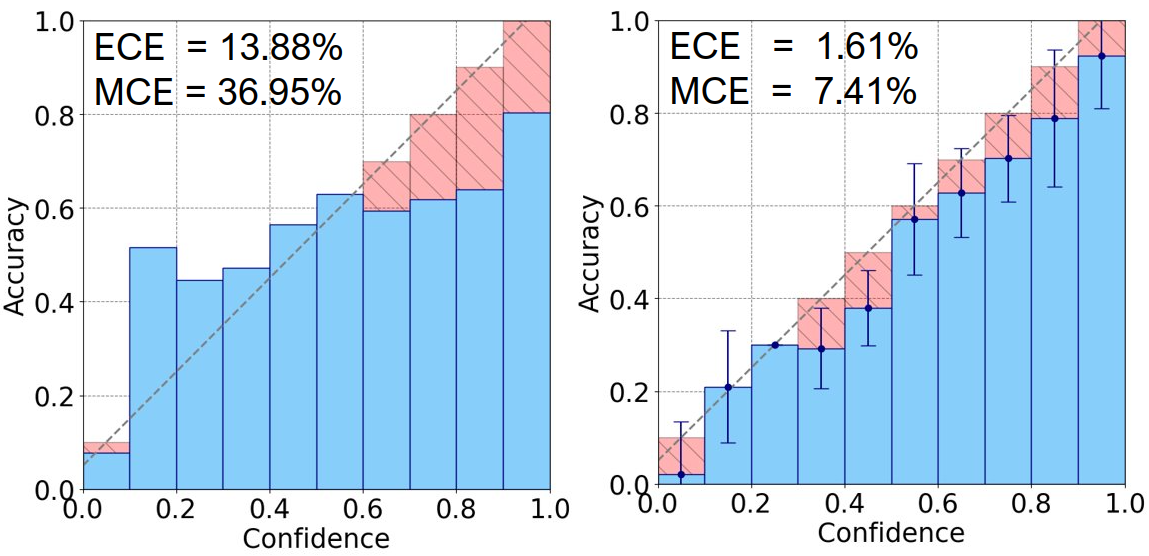} 
\caption{Calibration of a monolithic predictor (CNN, csp.) for racing car with horizon $k=100$. Left: uncalibrated, right: calibrated w/ isotonic regression and conformal bounds for $\alpha=0.05$.}
\label{fig:reliability-diagrams}
\end{figure}

An example reliability diagram (Fig.~\ref{fig:reliability-diagrams}), shows a common trend we witnessed among uncalibrated chance predictors: underconfident for the rejected class (below $0.5$), overconfident for the chosen class (above $0.5$). More results are shown in Tab.~\ref{tab:ececar} and ~\ref{tab:ecepole}. We can see that our calibration reduces the overconfidence and leads to a lower ECE --- even for long prediction horizons, as per Tab.~\ref{tab:ececar} and~\ref{tab:ecepole}. The most effective calibrators were the isotonic regression for the racing car and ENIR for the cart pole. Furthermore, our calibrated predictive intervals are well-aligned with the calibration uncertainty. This leads us to conclude that safety chance prediction is a more suitable formulation for highly uncertain image-controlled autonomous systems.

\begin{figure}[t]
\centering
\includegraphics[width=0.85\columnwidth]{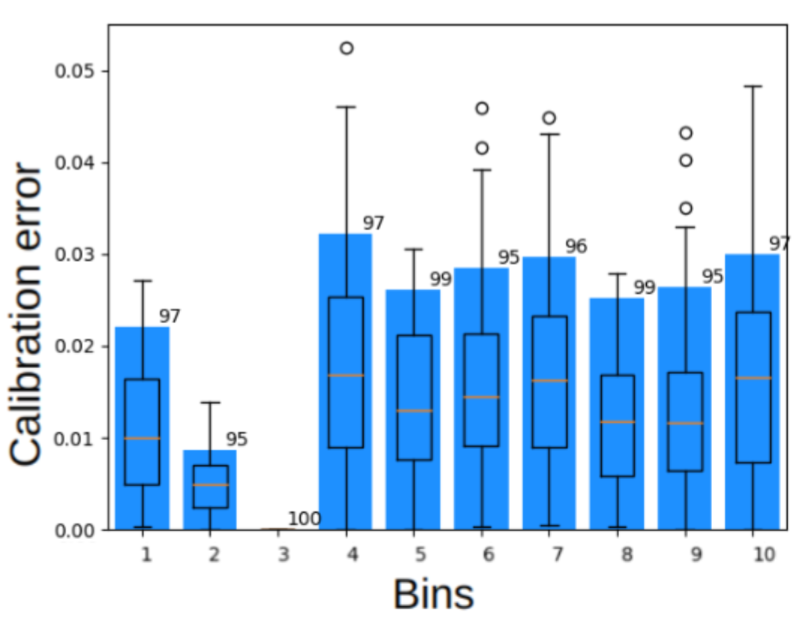} 

\caption{
Our conformal bounds of monolithic predictor (CNN, csp.) for cart pole with horizon $k=18$ for $\alpha=0.05$ (blue) contain more than 95\% of the true calibration errors (box and whisker plots).
}
\label{fig:coverage}
\end{figure}

\paragraph{Result 5: Conformal calibration coverage is reliable.} 
Supporting our theoretical claims, the predicted intervals for calibration errors contained the true error values from the test data (Fig. ~\ref{fig:coverage}). The sizes of these intervals are in Tab.~\ref{tab:intervalcar} and ~\ref{tab:intervalpole}, where the coverage of racing car is $97.74\%\pm3.46\%$ and cart pole is $95.91\%\pm6.65\%$. 
Our average error bound is $ 0.0924\pm0.0438$ for the racing car and $ 0.0409\pm0.0200$ for the cartpole. 
Therefore, our predicted chance $\pm$ its bin's calibration bound can be used reliably and informatively online.  

\begin{table*}[th]    
\caption{ECE for the racing car: before (white rows) and after (gray rows), for horizons 20 to 200 with selected calibrators by hyperparameter-
tuning.}
    \centering
        \tiny 

    \begin{tabular}{llllllllllll}
        \toprule
       Predictor & Model & $k=20$ & $k=40$ & $k=60$ & $k=80$ & $k=100$ & $k=120$ & $k=140$ & $k=160$ & $k=180$ & $k=200$\\

        \midrule

mono. csp.& lat-LSTM
&.0252$\pm$.0443  &.0713$\pm$.0759  &.1133$\pm$.0872  &.1539$\pm$.1143  &.1865$\pm$.1270  &.1996$\pm$.1320  &.1982$\pm$.1245  &.2163$\pm$.1254  &.2246$\pm$.1363  &.2506$\pm$.1363

\\
 \rowcolor{gray!20}
mono. csp.& lat-LSTM
&.0030$\pm$.0056  &.0069$\pm$.0063  &.0109$\pm$.0134  &.0099$\pm$.0102  &.0125$\pm$.0157  &.0118$\pm$.0147  &.0141$\pm$.0107  &.0146$\pm$.0117  &.0153$\pm$.0145  &.0179$\pm$.0113

\\
mono. ind.& lat-LSTM
&.0230$\pm$.0418  &.0675$\pm$.0893  &.1197$\pm$.0968  &.1523$\pm$.1068  &.1608$\pm$.1142  &.1868$\pm$.1318  &.1592$\pm$.1191  &.1592$\pm$.1099  &.1675$\pm$.1226  &.1609$\pm$.1170

\\
 \rowcolor{gray!20}
 mono. ind.& lat-LSTM

&.0021$\pm$.0051  &.0054$\pm$.0095  &.0054$\pm$.0081  &.0070$\pm$.0081  &.0074$\pm$.0069  &.0100$\pm$.0105  &.0141$\pm$.0129  &.0125$\pm$.0121  &.0119$\pm$.0096  &.0127$\pm$.0098

\\
   mono. csp.& CNN&.0361$\pm$.0759  &.0511$\pm$.0790  &.0753$\pm$.0822  &.0754$\pm$.0872  &.1162$\pm$.0838  &.1225$\pm$.1016  &.1287$\pm$.1095  &.1659$\pm$.0983  &.1513$\pm$.0897  &.1506$\pm$.1098
\\
 \rowcolor{gray!20}

mono. csp.& CNN
&.0025$\pm$.0047  &.0037$\pm$.0054  &.0078$\pm$.0072  &.0061$\pm$.0100  &.0119$\pm$.0134  &.0067$\pm$.0077  &.0122$\pm$.0108  &.0140$\pm$.0148  &.0107$\pm$.0072  &.0121$\pm$.0123

\\
mono. ind.& CNN 
&.0311$\pm$.0755  &.0465$\pm$.0919  &.0715$\pm$.0973  &.0832$\pm$.0982  &.1034$\pm$.0941  &.1349$\pm$.1021  &.1262$\pm$.1125  &.1403$\pm$.1082  &.1390$\pm$.1129  &.1704$\pm$.1238

\\
 \rowcolor{gray!20}

mono. ind.& CNN
&.0037$\pm$.0083  &.0028$\pm$.0060  &.0079$\pm$.0099  &.0076$\pm$.0109  &.0054$\pm$.0069  &.0094$\pm$.0093  &.0096$\pm$.0092  &.0109$\pm$.0137  &.0090$\pm$.0097  &.0137$\pm$.0127

\\
comp. csp. & lat-LSTM

&.0017$\pm$.0045  &.0045$\pm$.0136  &.0040$\pm$.0125  &.0079$\pm$.0237  &.0048$\pm$.0164  &.0040$\pm$.0175  &.0068$\pm$.0273  &.0091$\pm$.0337  &.0064$\pm$.0218  &.0103$\pm$.0386

\\
 \rowcolor{gray!20}

comp. csp. & lat-LSTM
&\textbf{.0005$\pm$.0011}  &.0011$\pm$.0039  &.0007$\pm$.0027  &.0010$\pm$.0042  &.0010$\pm$.0051  &.0009$\pm$.0034  &.0010$\pm$.0036  &.0009$\pm$.0037  &.0011$\pm$.0036  &.0010$\pm$.0031

\\
comp. ind. & lat-LSTM 
&.0046$\pm$.0179  &.0020$\pm$.0073  &.0052$\pm$.0270  &.0055$\pm$.0285  &.0011$\pm$.0036  &.0012$\pm$.0052  &.0016$\pm$.0051  &.0015$\pm$.0053  &.0039$\pm$.0127  &.0036$\pm$.0128

\\
 \rowcolor{gray!20}

comp. ind. & lat-LSTM
&.0006$\pm$.0018  &\textbf{.0006$\pm$.0024}  &\textbf{.0002$\pm$.0008}  &\textbf{.0002$\pm$.0007 } &\textbf{.0001$\pm$.0005 } &\textbf{.0002$\pm$.0006  }&\textbf{.0007$\pm$.0022  }&\textbf{.0004$\pm$.0012  }&\textbf{.0007$\pm$.0022}  &\textbf{.0008$\pm$.0035}
\\

\\
        \bottomrule
    \end{tabular}  
    \label{tab:ececar}
\end{table*}
\begin{table*}[th]
    \caption{ECE for the cartpole: before (white rows) and after (gray rows), for horizons from 3 to 27 with selected calibrators by hyperparameter-
tuning.}
    \centering
    \tiny 
    \begin{tabular}{llllllllllll}
        \toprule
        
        Predictor & Model & $k=3$ & $k=6$ & $k=9$ & $k=12$ & $k=15$ & $k=18$ & $k=21$ & $k=24$ & $k=27$
\\

        \midrule

    mono. csp. & lat-LSTM  
&.0042$\pm$.0107  &.0034$\pm$.0079  &.0033$\pm$.0115  &.0018$\pm$.0038  &.0024$\pm$.0049  &.0098$\pm$.0213  &.0474$\pm$.1037  &.0760$\pm$.1662  &.1600$\pm$.2061

\\
       \rowcolor{gray!20} 
        mono. csp. & lat-LSTM  
&\textbf{.0007$\pm$.0011}  &\textbf{.0005$\pm$.0007}  &.0009$\pm$.0033  &.0008$\pm$.0017  &.0010$\pm$.0016  &.0014$\pm$.0028  &\textbf{.0031$\pm$.0060}  &\textbf{.0021$\pm$.0032 } &\textbf{.0073$\pm$.0075}

\\
 mono. ind. & lat-LSTM

&.0035$\pm$.0117  &.0043$\pm$.0110  &.0019$\pm$.0043  &.0020$\pm$.0037  &.0028$\pm$.0065  &.0096$\pm$.0225  &.0506$\pm$.1082  &.0776$\pm$.1688  &.1627$\pm$.2068

\\

    \rowcolor{gray!20} 
        mono. ind. & lat-LSTM  
&.0007$\pm$.0015  &.0010$\pm$.0023  &\textbf{.0006$\pm$.0009 } &\textbf{.0006$\pm$.0010}  &\textbf{.0009$\pm$.0017}  &\textbf{.0012$\pm$.0022}  &.0031$\pm$.0065  &.0039$\pm$.0106  &.0075$\pm$.0094

\\

mono. csp. & CNN &.4919$\pm$.4892  &.4915$\pm$.4886  &.4928$\pm$.4883  &.4919$\pm$.4847  &.4923$\pm$.4730  &.4930$\pm$.4431  &.4942$\pm$.4003  &.4919$\pm$.3652  &.4915$\pm$.3398

\\

        \rowcolor{gray!20} 
        mono. csp. & CNN  &.0009$\pm$.0023  &.0012$\pm$.0028  &.0012$\pm$.0026  &.0011$\pm$.0022  &.0020$\pm$.0038  &.0049$\pm$.0067  &.0056$\pm$.0078  &.0083$\pm$.0098  &.0062$\pm$.0071
\\
        mono. ind. & CNN  
&.4933$\pm$.4842  &.4923$\pm$.4859  &.4917$\pm$.4841  &.4924$\pm$.4856  &.4916$\pm$.4720  &.4926$\pm$.4428  &.4918$\pm$.3994  &.4928$\pm$.3653  &.4916$\pm$.3398
\\

       \rowcolor{gray!20} 
        mono. ind. & CNN &.0018$\pm$.0031  &.0012$\pm$.0021  &.0006$\pm$.0012  &.0015$\pm$.0033  &.0018$\pm$.0032  &.0047$\pm$.0075  &.0058$\pm$.0093  &.0077$\pm$.0083  &.0085$\pm$.0083
\\

        comp. csp. & lat-LSTM 
&.0389$\pm$.0619  &.0631$\pm$.0651  &.1029$\pm$.0964  &.1378$\pm$.1626  &.1548$\pm$.1877  &.1492$\pm$.1637  &.1773$\pm$.1708  &.2140$\pm$.1891  &.2197$\pm$.2068

\\
    \rowcolor{gray!20} 
        comp. csp. & lat-LSTM 
 
&.0032$\pm$.0035  &.0044$\pm$.0057  &.0090$\pm$.0117  &.0068$\pm$.0074  &.0118$\pm$.0125  &.0106$\pm$.0115  &.0114$\pm$.0096  &.0123$\pm$.0145  &.0122$\pm$.0125

        \\
        comp. ind. & lat-LSTM 
&.0414$\pm$.0684  &.0547$\pm$.0888  &.0751$\pm$.0962  &.0730$\pm$.0646  &.0788$\pm$.0613  &.1003$\pm$.0837  &.1342$\pm$.1128  &.1492$\pm$.1277  &.2116$\pm$.1558 
\\
    \rowcolor{gray!20} 
        comp. ind. & lat-LSTM 
&.0049$\pm$.0106  &.0039$\pm$.0065  &.0045$\pm$.0090  &.0077$\pm$.0085  &.0076$\pm$.0078  &.0088$\pm$.0079  &.0092$\pm$.0077  &.0102$\pm$.0099  &.0119$\pm$.0107

\\
        \bottomrule
    \end{tabular}
    \label{tab:ecepole}
\end{table*}

\begin{table*}[th]    
\caption{Widths of conformal intervals for the racing car.}
    \centering
        \tiny 

    \begin{tabular}{llllllllllll}
        \toprule
       Predictor & Model & $k=20$ & $k=40$ & $k=60$ & $k=80$ & $k=100$ & $k=120$ & $k=140$ & $k=160$ & $k=180$ & $k=200$\\
        \midrule

        mono. csp. & lat-LSTM  &.0051$\pm$.0042  &.0232$\pm$.0270  &.0249$\pm$.0391  &.0222$\pm$.0187  &.0194$\pm$.0156  &.0178$\pm$.0187  &.0187$\pm$.0110  &.0194$\pm$.0231  &.0187$\pm$.0099  &.0169$\pm$.0100
 \\
      
        mono. ind. & lat-LSTM &\textbf{.0034$\pm$.0034 } &.0190$\pm$.0199  &.0137$\pm$.0130  &.0149$\pm$.0099  &.0120$\pm$.0100  &.0137$\pm$.0100  &\textbf{.0141$\pm$.0085 } &\textbf{.0137$\pm$.0099 } &\textbf{.0133$\pm$.0090 } &.0148$\pm$.0100
\\

        mono. csp. & CNN       &.0079$\pm$.0075  &.0267$\pm$.0283  &.0261$\pm$.0334  &\textbf{.0118$\pm$.0115}  &.0169$\pm$.0177  &\textbf{.0129$\pm$.0108 } &.0147$\pm$.0086  &.0138$\pm$.0101  &.0136$\pm$.0108  &.0168$\pm$.0104
\\
        mono. ind. & CNN     &.0086$\pm$.0072  &\textbf{.0114$\pm$.0169 } &\textbf{.0118$\pm$.0386 } &.0253$\pm$.0284  &\textbf{.0113$\pm$.0063 } &.0349$\pm$.0391  &.0223$\pm$.0188  &.0146$\pm$.0094  &.0174$\pm$.0105  &\textbf{.0118$\pm$.0058}
\\
        comp. csp. & lat-LSTM  &.0299$\pm$.0147  &.1360$\pm$.0572  &.2162$\pm$.0793  &.2659$\pm$.0896  &.2939$\pm$.1012  &.3183$\pm$.1095  &.3397$\pm$.1178  &.3554$\pm$.1223  &.3841$\pm$.1320  &.4126$\pm$.1420
 \\
        comp. ind. &lat-LSTM  &.0313$\pm$.0160  &.1079$\pm$.0500  &.1321$\pm$.0616  &.1623$\pm$.0680  &.2060$\pm$.0869  &.2393$\pm$.1034  &.2730$\pm$.1170  &.3047$\pm$.1339  &.3279$\pm$.1350  &.3610$\pm$.1395
 
\\
        \bottomrule
    \end{tabular}  
    \label{tab:intervalcar}
\end{table*}

\begin{table*}[th]    
\caption{Widths of conformal intervals for the cartpole.}
    \centering
        \tiny 

    \begin{tabular}{llllllllllll}
        \toprule
        Predictor & Model & $k=3$ & $k=6$ & $k=9$ & $k=12$ & $k=15$ & $k=18$ & $k=21$ & $k=24$ & $k=27$ \\
        \midrule
 
mono. csp. & lat-LSTM  &.0064$\pm$.0088  &.0059$\pm$.0082  &.0061$\pm$.0065  &\textbf{.0069$\pm$.0092}  &\textbf{.0097$\pm$.0093}  &\textbf{.0226$\pm$.0180}  &.0581$\pm$.0486  &.1270$\pm$.0922  &.0962$\pm$.0850

\\
mono. ind. & lat-LSTM & .0077$\pm$.0094  &.0059$\pm$.0095  &\textbf{.0038$\pm$.0062 } &.0074$\pm$.0092  &.0127$\pm$.0129  &.0234$\pm$.0171  &\textbf{.0573$\pm$.0472 } &.1258$\pm$.0911  &.0989$\pm$.0858

\\
mono. csp. &CNN&\textbf{.0055$\pm$.0081}  &\textbf{.0049$\pm$.0088}  &.0074$\pm$.0127  &.0097$\pm$.0170  &.0200$\pm$.0235  &.0347$\pm$.0310  &.0449$\pm$.0298  &.0599$\pm$.0319  &.0694$\pm$.0273

\\
mono. ind. & CNN &.0134$\pm$.0123  &.0102$\pm$.0127  &.0106$\pm$.0168  &.0091$\pm$.0139  &.0190$\pm$.0211  &.0330$\pm$.0283  &.0470$\pm$.0311  &\textbf{.0588$\pm$.0295}  &\textbf{.0676$\pm$.0258}

\\
   comp. csp. & lat-LSTM  &.0275$\pm$.0243  &.0464$\pm$.0312  &.0525$\pm$.0324  &.0584$\pm$.0340  &.0579$\pm$.0319  &.0625$\pm$.0282  &.0721$\pm$.0298  &.0734$\pm$.0233  &.0754$\pm$.0263

\\

comp. ind. & lat-LSTM &.0249$\pm$.0248  &.0306$\pm$.0304  &.0426$\pm$.0325  &.0500$\pm$.0291  &.0536$\pm$.0301  &.0601$\pm$.0312  &.0647$\pm$.0262  &.0678$\pm$.0285  &.0803$\pm$.0273

\\
        \bottomrule
    \end{tabular}  
    \label{tab:intervalpole}
\end{table*}

\section{Related Work}\label{sec:relwork}


\paragraph{Performance and safety evaluation for autonomy}

A variety of recent research enables autonomous systems to \emph{self-evaluate} their 
 competency~\cite{basich_competence-aware_2022}. Performance metrics vary significantly and include such examples as the time to navigate to a goal location or whether a safety constraint was violated, which is the focus of our paper. For systems that have access to their low-dimensional state, such as the velocity and the distance from the goal, \emph{hand-crafted indicators} have been successful in measuring performance degradation. These indicators have been referred to  robot vitals~\cite{9834068}, alignment checkers~\cite{9812030}, assumption monitors~\cite{ruchkin_confidence_2022}, and operator trust~\cite{conlon_im_2022}. Typically, these indicators rely on domain knowledge and careful offline analysis of the system, which are difficult to obtain and perform for high-dimensional systems. 

Autonomy with deep neural network (DNN) controllers is vulnerable to distribution shift  \cite{MORENOTORRES2012521,HUANG2020100270} and difficult to analyze. On the model-based side, a number of \emph{closed-loop verification approaches} perform analyze reachability and safety of a DNN-controlled system~\cite{ivanov_verisig_2021,tran_verification_2022}, but when applied to vision-based systems, require detailed modeling of the vision subsystem and have limited scalability~\cite{santa_cruz_nnlander-verif_2022,hsieh_verifying_2022}. On the other hand, \emph{model-free safety predictions} rely on the correlation between performance and uncertainty measures/anomaly scores, such as autoencoder reconstruction errors~\cite{9284027} and distances to representative training data~\cite{yang_interpretable_2022};  
however, these approaches fail to fully utilize the information about the system's closed-loop nature. Striking the balance between model-based and model-free approaches are \emph{black-box statistical methods} for risk assessment and safety verification \cite{cleaveland_risk_2022,10.1145/3365365.3382209,9797578,michelmore_uncertainty_2020}. They usually require low-dimensional states and rich outcome labels (such as signal robustness) --- an assumption that we are relaxing in this work while addressing a similar problem. 
 




\paragraph{Trajectory prediction}

A common way of predicting the system's performance and safety is by inferring it from \emph{predicted trajectories}. 
Classic approaches consider model-based prediction~\cite{lefevre_survey_2014}, for instance, estimating collision risk with bicycle dynamics and Kalman filtering~\cite{5284727}. One common approach with safety guarantees is \emph{Hamilton-Jacobi} (HJ) reachability~\cite{li_prediction-based_2021,nakamura2023online}, which requires precomputation based on a dynamical model. 
Among many deep learning-based predictors~\cite{huang_survey_2022}, a recently popular architecture is \emph{Trajectron++}~\cite{salzmann_trajectron_2020} takes in high-dimensional scene graphs and outputs future trajectories for multiple agents. A conditional VAE~\cite{NIPS2015_8d55a249} is adopted in Trajectron++ to add constraints at the decoding stage. Learning-based trajectory predictions can be augmented with conformal prediction to improve their reliability~\cite{lindemann_conformal_2023,muthali_multi-agent_2023}. 
In comparison, our work eschews handcrafted scene and state representations, instead using image and latent representations that are informed by safety.
\paragraph{Safe control}
Safe planning and control solve a  problem complementary to ours: controlling a system safely, usually with respect to a dynamical model, such as motion planning under uncertainty for uncertain  systems~\cite{knuth_planning_2021,hibbard_safely_2022,chou_synthesizing_2023} and a growing body of work on control barrier functions~\cite{ames_control_2019,xiao_safe_2023}. Some recent approaches are robust to measurement errors induced by learning-based perception~\cite{dean_guaranteeing_2021,yang_safe_2023}, but at their core, they still hinge on the utilization of low-dimensional states, a paradigm we aim to circumvent entirely.  Importantly, one of our constraints is assuming the existence of an end-to-end controller (possibly implemented with the above methods), with no room for modifications.
\paragraph{Confidence calibration}
The softmax scores of classification neural networks can be interpreted as probabilities for each class; however, standard training leads to miscalibrated neural networks~\cite{guo_calibration_2017,minderer_revisiting_2021}, as measured by \emph{Brier score} and \emph{Expected Calibration Error} (ECE). Calibration approaches can be categorized into \emph{extrinsic (post-hoc) calibration} added on top of a trained network, such as Platt and temperature scaling~\cite{platt_probabilistic_1999,guo_calibration_2017}, isotonic regression~\cite{zadrozny_transforming_2002}, and histogram/Bayesian binning~\cite{naeini_obtaining_2015} --- and \emph{intrinsic calibration} to modify the training, such as ensembles~\cite{zhang_mix-n-match_2020}, adversarial training~\cite{lee_training_2018}, and learning from hints~\cite{devries_learning_2018}, error distances~\cite{xing_distance-based_2019}, or true class probabilities~\cite{corbiere_addressing_2019}. Similar techniques have been developed for calibrated regression~\cite{vovk_conformal_2020,marx_modular_2022}. It is feasible to obtain calibration guarantees for safety chance predictions~\cite{ruchkin_confidence_2022,cleaveland_conservative_2023}, but it requires a low-dimensional model-based  setting. To the authors' best knowledge, such guarantees have not been instantiated in a model-free autonomy setting.

\paragraph{Deep surrogate models}

For systems with high-resolution images and complex dynamical models, trajectory predictions become particularly challenging.  Physics-based methods and classical machine learning methods may not be applicable to these tasks, leading to the almost exclusive application of deep learning for prediction. \emph{Sequence prediction models}, which originate in deep video prediction~\cite{Oprea_2022}, usually do not perform well since they require long horizons and a large number of samples from each controller. 
However, incorporating additional information, such as states, observations, and actions, in sequence predictions can improve their performance~\cite{strickland2018deep}, and our predictors take advantage of that.

High-dimensional sensor data contains substantial redundant and irrelevant information, which leads to higher computational costs. 
Generative adversarial networks (GANs) map the observations into low-dimensional latent space, which can enable conventional assurance tools~\cite{katz2021verification}. 
Low-dimensional representations can incorporate conformal predictions to monitor the real-time performance~\cite{boursinos_assurance_2021}. \emph{World models}~\cite{ha_recurrent_2018} are used as a surrogate method of training a controller instead of using reinforcement learning (RL). They achieve better results to train the controllers than the basic reinforcement learning method. This kind of model can learn how the dynamical system works, which collects random observations and actions as inputs and outputs the future states.   Then it uses a VAE to compress observations into a latent space (low-dimensional vectors) and trains a mixed-density recurrent neural network (MDN-RNN) to forecast the next-step latent observations. The decoder of the VAE maps the latent vector into observation space.  This model is adopted to do trajectory predictions and get the competency assessment through the forecasting trajectory~\cite{acharya_competency_2022}. The original world model uses a VAE to compress the observations and an MDN-RNN to perform sequence prediction. Lately, more advanced models like \emph{DreamerV2}~\cite{hafner2022mastering} and \emph{IRIS}~\cite{micheli2023transformers} adopt recent deep computer vision models like vision transformers to achieve more vivid images. The trajectory predicted by world models can also provide uncertainty measurement for agents~\cite{acharya2023learning}. In our method, we adopt the architecture of the initial version of world models which have VAE and recurrent models and compare them with other deep learning models like the recurrent network alone and convolutional recurrent networks. On top of these approaches, we have developed chance calibration guarantees, which were previously unexplored.






\section{Discussion and Conclusion}
\label{sec:conclusion}

\emph{Limitations:} our predictor family's scope is limited to systems where safety can be inferred from raw sensor data. Also, our safety evaluators are system-specific, and developing them may require substantial effort; however, this effort is balanced with the savings of not needing to design state representations and develop dynamical models (both of which would be system-specific as well).  Besides, obtaining negative safety samples can have a high cost in reality, which can be overcome with transfer learning from simulation and generative models.


Some approaches have performed poorly (not reported due to space limits). One such approach is \emph{flexible-horizon predictors}, which output the time to the first expected safety violation. Their prediction space appeared too complex and insufficiently regularized, leading to poor performance. Also, \emph{latent space evaluators} (as opposed to image evaluators) have shown particularly poor performance, which indicates that the latent state vectors failed to preserve safety information. Thus, learning strong safety-informed representations for autonomy remains an open problem. Though quantized latent expression now is used for video predictions ~\cite{walker2021predicting} and video generation~\cite{yan2021videogpt}, the \emph{CLIP}~\cite{radford2021learning}, a multimodal unsupervised method, unifies the representations of images and texts.   With the rise of large language models, meaning representations in multimodal autoregressive models~\cite{liu2023meaning} may provide more possibilities to explore the surrogate dynamics of autonomy systems.

Future work includes adding physical constraints to latent states similar to Neural ODEs~\cite{wen_social_2022}, jointly learning forecasters and evaluators to overcome distribution shifts, using world-model transformer architectures, and applying predictors to physical systems.

\section{Acknowledgments}

We would like to thank Kaleb Smith (NVIDIA AI Technology Center), Yuang Geng (University of Florida), and anonymous reviewers for their insightful feedback on this research.



\bibliography{new,response,ivans-bibs/full-autogenerated}

\newpage

\section*{Appendix
}

\subsection{A \quad Hyperparameters}
 We used Pytorch 1.13.1 with the Adam optimizer. The maximum training epoch is 500 for VAEs and 100 for predictors. The starting learning rate of $10^{-3}$, and the reduction factor of 0.1 with a patience of 5 epochs. The early-stopping patience is 15 epochs. The batch size of the classifier, the latent predictor, and the image predictor are 128, 64, and 16. The architecture for the evaluator and monolithic single-image predictor consists of 2 convolutional layers with kernel sizes of 3 and 5 respectively, 3 linear layers and 1 softmax layer. Each 2D convolution layer is combined with max-pooling layers with the size of 2*2 with stride 2. For the VAE, the latent size is 32. The encoder has 4 convolutional layers and 2 linear layers. Each convolutional layer is combined with a ReLU, and the decoder has a similar structure with transposed convolutional layers. The structure of the monolithic sequence predictor and the composite latent predictor is one LSTM layer and one linear layer. The monolithic predictor has an extra softmax layer, and has an output size of 2, while the composite has an output size of 32. The safety loss uses $\lambda_1 =1$ and $ \lambda_2 = 4096$, which equals the total pixel count in our images. The miscoverage level is $\alpha=0.05$. 
 
 The following tables contain the hyperparameters: 
 \begin{itemize}
     \item The dataset size and prediction horizons are shown (Tab.~\ref{tab:hyperparams})
     \item The architecture of our evaluator and single-image predictors (Tab.~\ref{tab:architecture-cnn})
     \item The architecture of our image forecaster (Tab.~\ref{tab:architecture-conv-lstm})
     \item The architecture of our latent LSTM, which uses an encoder to compress images (Tab.~\ref{tab:architecture-latent-lstm})
     \item The architecture of our latent forecaster (Tab.~\ref{tab:architecture-comp-lat})
     \item The architecture of our VAEs (Tab.~\ref{tab:architecture-vae})
 \end{itemize}



\subsection{B  \quad Result Tables}

\begin{table*}[th]    
\caption{Predictors' F1 scores for the racing car.}
    \centering
        \tiny 

    \begin{tabular}{llllllllllll}
        \toprule
    Predictor & Model & $k=20$ & $k=40$ & $k=60$ & $k=80$ & $k=100$ & $k=120$ & $k=140$ & $k=160$ & $k=180$ & $k=200$

 \\
        \midrule


mon. csp. &lat-LSTM &\textbf{96.7 $\pm$ 0.5 }&\textbf{93.1 $\pm$ 1.2} &\textbf{87.8 $\pm$ 2.2} &\textbf{83.4 $\pm$ 2.6} &\textbf{80.9 $\pm$ 3.0 }&\textbf{79.2 $\pm$ 2.5}&77.5 $\pm$ 2.1 &7\textbf{5.5 $\pm$ 2.8 }&\textbf{74.9 $\pm$ 6.0 }&72.3 $\pm$ 3.1

\\

mon. ind. &lat-LSTM &96.6 $\pm$ 0.6 &89.9 $\pm$ 2.1 &86.2 $\pm$ 2.9 &82.4 $\pm$ 4.8 &80.8 $\pm$ 4.7 &78.9 $\pm$ 5.5 &\textbf{77.6 $\pm$ 5.1} &72.9 $\pm$ 8.9 &70.7 $\pm$ 5.6 &69.7 $\pm$ 6.1
\\


 mon. csp. &CNN &92.3 $\pm$ 1.6 &87.3 $\pm$ 3.7 &83.3 $\pm$ 5.3 &80.9 $\pm$ 3.4 &78.4 $\pm$ 10.6 &77.3 $\pm$ 5.2 &75.5 $\pm$ 2.6 &74.0 $\pm$ 10.9 &71.4 $\pm$ 6.9 &72.0 $\pm$ 9.2

\\

 mon. ind. &CNN &92.6 $\pm$ 1.6 &87.9 $\pm$ 2.8 &82.4 $\pm$ 4.8 &78.9 $\pm$ 5.8 &77.6 $\pm$ 5.1 &72.9 $\pm$ 8.9 &70.7 $\pm$ 5.6 &69.7 $\pm$ 6.1 &69.1 $\pm$ 5.8 &68.6 $\pm$ 5.3

\\




 comp. csp. &lat-LSTM e&92.8 $\pm$ 0.5 &84.5 $\pm$ 1.7 &74.9 $\pm$ 7.7 &66.1 $\pm$ 12.0 &52.9 $\pm$ 17.7 &47.3 $\pm$ 20.8 &43.7 $\pm$ 21.6 &41.8$\pm$21.7&38.0 $\pm$ 23.8 &35.9 $\pm$ 25.8
\\
       \rowcolor{gray!20} 
 comp. csp. &md-LSTM e 
 &33.3$\pm$ 47.1
 &66.7$\pm$ 47.1
 &66.7$\pm$ 47.1 
 &66.6$\pm$ 47.1
 &99.9$\pm$ 0.1
 &99.9$\pm$ 0.1 
 &99.7$\pm$ 0.3
 &99.6$\pm$ 0.4
 &99.5$\pm$ 0.4
\\


 comp. csp. &lat-LSTM v &93.5 $\pm$ 0.8 &86.6 $\pm$ 0.4 &80.9 $\pm$ 1.6 &75.8 $\pm$ 4.8 &71.6 $\pm$ 8.1 &69.8 $\pm$ 8.0 &67.4 $\pm$ 6.3 &66.1 $\pm$ 6.0 &63.6 $\pm$ 4.7 &60.2 $\pm$ 3.5

\\

  comp. ind.  &lat-LSTM e&94.5 $\pm$ 0.1 &83.7 $\pm$ 3.3 &68.9 $\pm$ 12.1 &67.0 $\pm$ 11.8 &63.3 $\pm$ 13.9 &61.3 $\pm$ 15.7 &59.2 $\pm$ 16.4 &56.3 $\pm$ 17.2 &54.2 $\pm$ 17.0 &51.9 $\pm$ 16.5
\\

       \rowcolor{gray!20} 

  comp. ind.  &md-LSTM e
   &0.0$\pm$ 0.0 &0.0$\pm$ 0.0 &65.4$\pm$ 46.2 &50.9$\pm$ 36.1 &81.5$\pm$ 13.7 &84.3$\pm$ 12.4 &84.5$\pm$ 12.2 &80.7$\pm$ 15.3 &73.7$\pm$ 19.7
\\

comp. ind. &lat-LSTM v &93.4 $\pm$ 0.5 &83.9 $\pm$ 2.1 &74.3 $\pm$ 6.9 &70.6 $\pm$ 8.9 &71.9 $\pm$ 4.3 &68.5 $\pm$ 7.0 &64.6 $\pm$ 10.5 &62.8 $\pm$ 9.4 &59.9 $\pm$ 10.7 &56.3 $\pm$ 10.6

\\

comp. csp &conv-LSTM  e&6.7 $\pm$ 1.9 &6.8 $\pm$ 1.8 &6.7 $\pm$ 1.6 &6.6 $\pm$ 1.5 &6.4 $\pm$ 1.6 &6.3 $\pm$ 1.7 &6.2 $\pm$ 1.8 &6.0 $\pm$ 1.6 &5.7 $\pm$ 1.3 &4.9 $\pm$ 0.8

\\
comp. csp &conv-LSTM v1&37.8 $\pm$ 27.6 &37.1 $\pm$ 26.6 &36.3 $\pm$ 25.6 &35.8 $\pm$ 24.9 &35.3 $\pm$ 24.3 &34.7 $\pm$ 23.9 &34.2 $\pm$ 22.9 &33.6 $\pm$ 22.3 &33.3 $\pm$ 21.9 &32.2 $\pm$ 20.9

\\
comp. csp &conv-LSTM v2&4.4 $\pm$ 2.6 &4.3 $\pm$ 2.4 &4.2 $\pm$ 2.2 &4.1 $\pm$ 2.2 &4.0 $\pm$ 2.1 &3.9 $\pm$ 2.0 &3.9 $\pm$ 1.9 &3.8 $\pm$ 1.9 &3.8 $\pm$ 1.9 &3.7 $\pm$ 1.6

\\

        \bottomrule
    \end{tabular}  
    \label{tab:f1car}
\end{table*}

\begin{table*}[th]    
\caption{Predictors' FPRs for the racing car.}
    \centering
        \tiny 

    \begin{tabular}{llllllllllll}
        \toprule
    Predictor & Model & $k=20$ & $k=40$ & $k=60$ & $k=80$ & $k=100$ & $k=120$ & $k=140$ & $k=160$ & $k=180$ & $k=200$

 \\
        \midrule


mon. csp. & lat-LSTM &3.4 $\pm$ 0.6 &8.6 $\pm$ 2.1 &17.7 $\pm$ 3.5 &25.2 $\pm$ 5.8 &30.4 $\pm$ 5.1 &33.9 $\pm$ 6.4 &38.4 $\pm$ 6.5 &38.2 $\pm$ 5.6 &37.3 $\pm$ 11.7 &37.8 $\pm$ 7.4\\

mon. ind. & lat-LSTM&4.2 $\pm$ 1.7 &13.2 $\pm$ 5.6 &22.0 $\pm$ 8.2 &25.9 $\pm$ 9.4 &34.4 $\pm$ 11.2 &33.4 $\pm$ 8.4 &32.5 $\pm$ 6.6 &34.9 $\pm$ 7.1 &31.8 $\pm$ 4.9 &30.8 $\pm$ 4.1

\\

mon. csp. & CNN  &9.1    $\pm$ 1.6 &17.4 $\pm$ 3.7 &  24.6$\pm$5.3&32.2 $\pm$ 3.4 &36.9 $\pm$ 10.6 &36.3 $\pm$ 5.2 &38.8 $\pm$ 2.6 &39.0 $\pm$ 2.2 &43.7 $\pm$ 9.4 &40.4 $\pm$ 9.2

\\
mon. ind. & CNN &9.6 $\pm$ 4.2 &16.5 $\pm$ 7.2 &25.9 $\pm$ 9.2 &29.6 $\pm$ 10.4 &31.3 $\pm$ 9.9 &33.4 $\pm$ 8.4 &32.5 $\pm$ 6.6 &34.9 $\pm$ 7.1 &31.8 $\pm$ 4.9 &27.8 $\pm$ 4.1

\\



comp. csp. &lat-LSTM  e&11.1 $\pm$ 4.7 &23.0 $\pm$ 9.4 &29.7 $\pm$ 12.8 &28.9 $\pm$ 11.6 &22.8 $\pm$ 10.8 &21.0 $\pm$ 13.4 &20.3 $\pm$ 15.6 &19.8 $\pm$ 17.8 &18.7 $\pm$ 21.3 &18.2 $\pm$ 21.3

\\
       \rowcolor{gray!20} 

 comp. csp. &md-LSTM  e&100.0$\pm$ 0.0 &100.0$\pm$ 0.0 &100.0$\pm$ 0.0 &nan$\pm$ nan &nan$\pm$ nan &nan$\pm$ nan &nan$\pm$ nan &nan$\pm$ nan &nan$\pm$ nan 
\\

comp. csp. &lat-LSTM v&45.0 $\pm$ 17.0 &76.5 $\pm$ 14.4 &84.8 $\pm$ 10.3 &84.9 $\pm$ 11.4 &82.7 $\pm$ 13.6 &84.0 $\pm$ 13.0 &84.7 $\pm$ 12.5 &85.3 $\pm$ 12.2 &86.3 $\pm$ 10.3 &87.7 $\pm$ 10.5

\\





       \rowcolor{gray!20} 

 comp. ind. &md-LSTM &100.0$\pm$ 0.0 &100.0$\pm$ 0.0 &nan$\pm$ nan &nan$\pm$ nan &nan$\pm$ nan &nan$\pm$ nan &nan$\pm$ nan &nan$\pm$ nan &nan$\pm$ nan

 \\
comp. ind. &lat-LSTM e&35.3 $\pm$ 17.1 &64.2 $\pm$ 17.1 &72.7$\pm$15.6&73.6 $\pm$ 16.2 &72.5 $\pm$ 16.8 &72.4 $\pm$ 22.1 &70.9 $\pm$ 27.1 &69.3 $\pm$ 30.5 &68.4 $\pm$ 33.3 &68.3 $\pm$ 34.9

\\

comp. ind. &lat-LSTM v&61.8 $\pm$ 18.8 &78.2 $\pm$ 12.1 &79.9 $\pm$ 11.3 &79.2 $\pm$ 11.8 &80.8 $\pm$ 11.2 &79.0 $\pm$ 20.0 &76.9 $\pm$ 21.0 &77.5 $\pm$ 22.3 &75.1 $\pm$ 25.9 &75.7 $\pm$ 26.3


 \\

comp. csp &conv-LSTM e&\textbf{0.6 $\pm$ 0.5} &\textbf{1.1 $\pm$ 0.9} &\textbf{1.4 $\pm$ 1.1} &\textbf{1.7 $\pm$ 1.1 }&1.9$\pm$ 1.2&2.0 $\pm$ 1.2 &2.2 $\pm$ 1.1 &2.4 $\pm$ 1.2 &2.6 $\pm$ 1.4 &3.0 $\pm$ 1.5

\\

comp. csp &conv-LSTM v1&16.2 $\pm$ 10.3 &19.9 $\pm$ 13.7 &23.4 $\pm$ 16.0 &28.4 $\pm$ 17.1 &32.2 $\pm$ 17.7 &37.4 $\pm$ 18.3 &43.2 $\pm$ 19.3 &47.8 $\pm$ 19.5 &11.0 $\pm$ 7.9 &26.1 $\pm$ 16.6

\\

comp. csp &conv-LSTM v2&1.1 $\pm$ 0.4 &1.4 $\pm$ 0.8 &1.6 $\pm$ 1.0 &1.9 $\pm$ 1.1 &\textbf{1.8 $\pm$ 1.1} &\textbf{1.9 $\pm$ 1.1} &\textbf{2.0 $\pm$ 1.2 }&\textbf{2.0 $\pm$ 1.2} &\textbf{2.2 $\pm$ 1.4} &\textbf{2.2 $\pm$ 1.4}

 \\

        \bottomrule
    \end{tabular}  
    \label{tab:fprcar}
\end{table*}

\begin{table*}[th]    
\caption{Predictors' F1 scores for the cartpole.}
    \centering
        \scriptsize 

    \begin{tabular}{llllllllllll}
        \toprule
        Predictor & Model & $k=3$ & $k=6$ & $k=9$ & $k=12$ & $k=15$ & $k=18$ & $k=21$ & $k=24$ & $k=27$ \\
        \midrule

mono. csp. &lat-LSTM 

&99.8 $\pm$ 0.1 &99.8 $\pm$ 0.1 &99.6 $\pm$ 0.2 &99.5 $\pm$ 0.2 &99.4 $\pm$ 0.3 &\textbf{99.3 $\pm$ 0.3} &\textbf{98.4 $\pm$ 0.8 }&\textbf{93.4 $\pm$ 6.7} &\textbf{86.7 $\pm$ 13.9 }
 \\

mono. ind. &lat-LSTM 
&\textbf{99.8 $\pm$ 0.0} &\textbf{99.8 $\pm$ 0.0} &\textbf{99.8 $\pm$ 0.0 }&\textbf{99.8 $\pm$ 0.0 }&\textbf{99.5 $\pm$ 0.1 }&98.7 $\pm$ 0.7 &93.8 $\pm$ 6.4 &88.1 $\pm$ 12.2 &85.3 $\pm$ 13.4  \\

mono. sp. &CNN 
&99.8 $\pm$ 0.1 &99.8 $\pm$ 0.1 &99.7 $\pm$ 0.1 &99.4 $\pm$ 0.1 &98.3 $\pm$ 1.1 &94.8 $\pm$ 5.4 &90.6 $\pm$ 9.7 &85.4 $\pm$ 15.4 &81.8 $\pm$ 18.6  \\

mono. ind. &CNN 

&99.5 $\pm$ 0.2 &99.6 $\pm$ 0.2 &99.5 $\pm$ 0.1 &99.3 $\pm$ 0.1 &98.3 $\pm$ 0.7 &94.8 $\pm$ 4.5 &90.6 $\pm$ 8.3 &86.9 $\pm$ 11.2 &84.7 $\pm$ 12.1 \\

comp. csp. &lat-LSTM e&92.9 $\pm$ 0.6 &90.4 $\pm$ 4.3 &60.5 $\pm$ 31.5 &29.3 $\pm$ 40.5 &43.3 $\pm$ 36.1 &33.1 $\pm$ 23.5 &22.2 $\pm$ 28.6 &19.8 $\pm$ 26.9 &21.9 $\pm$ 29.0 \ \\

       \rowcolor{gray!20} 

 comp. csp. &md-LSTM e &92.6$\pm$ 1.0 &85.3$\pm$ 1.3 &90.1$\pm$ 1.5 &93.3$\pm$ 2.1 &91.4$\pm$ 2.4 &88.6$\pm$ 2.2 &85.6$\pm$ 2.9 &84.4$\pm$ 3.1 &82.7$\pm$ 2.6 &82.7$\pm$ 2.2

 \\

comp. csp. &lat-LSTM v&88.6 $\pm$ 4.4 &85.5 $\pm$ 7.2 &67.3 $\pm$ 12.1 &29.9 $\pm$ 30.1 &49.9 $\pm$ 34.2 &55.8 $\pm$ 19.8 &64.9 $\pm$ 8.5 &64.1 $\pm$ 5.8 &51.2 $\pm$ 13.7 
\\

comp. ind. &lat-LSTM e &94.8 $\pm$ 1.9 &93.1 $\pm$ 3.7 &83.7 $\pm$ 10.3 &48.1 $\pm$ 25.7 &40.4 $\pm$ 28.4 &60.2 $\pm$ 10.6 &66.2 $\pm$ 3.2 &62.8 $\pm$ 5.4 &62.6 $\pm$ 9.0 
\\

       \rowcolor{gray!20} 

  comp. ind. &md-LSTM e &69.7$\pm$ 26.9 &68.6$\pm$ 20.8 &73.8$\pm$ 15.8 &75.1$\pm$ 14.0 &75.2$\pm$ 12.7 &74.4$\pm$ 13.3 &73.3$\pm$ 13.3 &73.0$\pm$ 13.5 &72.8$\pm$ 11.3 &74.4$\pm$ 7.

\\

comp. ind. &lat-LSTM v&88.3 $\pm$ 4.1 &87.1 $\pm$ 6.4 &83.9 $\pm$ 9.7 &61.9 $\pm$ 24.2 &52.5 $\pm$ 22.9 &68.6 $\pm$ 8.4 &76.4 $\pm$ 3.1 &73.4 $\pm$ 1.6 &70.5 $\pm$ 2.0 
 
 \\

comp. ind. &conv-LSTM e &64.0 $\pm$ 1.8 &0.0 $\pm$ 0.0 &0.0 $\pm$ 0.0 &0.0 $\pm$ 0.0 &0.0 $\pm$ 0.0 &0.0 $\pm$ 0.0 &0.0 $\pm$ 0.0 &0.0 $\pm$ 0.0 &0.0 $\pm$ 0.0  \\

comp. ind. &conv-LSTM v&67.3 $\pm$ 2.9 &0.0 $\pm$ 0.0 &0.0 $\pm$ 0.0 &0.0 $\pm$ 0.0 &0.0 $\pm$ 0.0 &0.0 $\pm$ 0.0 &0.0 $\pm$ 0.0 &0.0 $\pm$ 0.0 &0.0 $\pm$ 0.0 
 \\
        \bottomrule
    \end{tabular}  
    \label{tab:f1pole}
\end{table*}

\begin{table*}[th]    
\caption{Predictors' FPR for the cart pole.}
    \centering
        \scriptsize 

    \begin{tabular}{llllllllllll}
        \toprule
        Predictor & Model & $k=3$ & $k=6$ & $k=9$ & $k=12$ & $k=15$ & $k=18$ & $k=21$ & $k=24$ & $k=27$ \\
        \midrule

mono. csp. &lat-LSTM &0.1 $\pm$ 0.0 &0.2 $\pm$ 0.1 &0.5 $\pm$ 0.5 &0.6 $\pm$ 0.4 &0.7 $\pm$ 0.5 &0.8 $\pm$ 0.5 &2.5 $\pm$ 1.8 &9.3 $\pm$ 8.8 &20.8 $\pm$ 20.4   \\

mono. ind. &lat-LSTM  &0.4 $\pm$ 0.1 &0.3 $\pm$ 0.1 &0.2 $\pm$ 0.1 &0.3 $\pm$ 0.0 &0.3 $\pm$ 0.2 &0.3 $\pm$ 0.2 &0.4 $\pm$ 0.2 &0.4 $\pm$ 0.3 &1.2 $\pm$ 0.8 \\

 mono. csp. &CNN 

&\textbf{0.0 $\pm$ 0.0} &\textbf{0.0 $\pm$ 0.0 }&\textbf{0.0 $\pm$ 0.0} &\textbf{0.0 $\pm$ 0.0 }&2.7 $\pm$ 2.6 &10.8 $\pm$ 11.4 &16.3 $\pm$ 17.1 &17.3 $\pm$ 15.1 &19.3 $\pm$ 13.6 
\\
mono. ind. &CNN 
&0.4 $\pm$ 0.1 &0.4 $\pm$ 0.1 &0.4 $\pm$ 0.1 &0.6 $\pm$ 0.2 &1.1 $\pm$ 0.5 &0.9 $\pm$ 0.7 &0.8 $\pm$ 0.6 &0.8 $\pm$ 0.5 &1.0 $\pm$ 0.3 \\

comp. csp. &lat-LSTM e &9.3 $\pm$ 0.5 &17.0 $\pm$ 4.0 &22.1 $\pm$ 4.9 &26.3 $\pm$ 11.0 &23.1 $\pm$ 8.0 &22.8 $\pm$ 6.4 &30.9 $\pm$ 9.4 &36.2 $\pm$ 12.7 &36.1 $\pm$ 12.2 \\

       \rowcolor{gray!20} 

comp. csp. &lat-LSTM e

 &40.8$\pm$ 28.0 &49.1$\pm$ 34.9 &52.9$\pm$ 38.0 &50.9$\pm$ 38.6 &51.3$\pm$ 34.1 &54.3$\pm$ 29.1 &59.6$\pm$ 26.1 &64.0$\pm$ 28.4 &69.6$\pm$ 34.4 &68.6$\pm$ 41.3

\\

comp. csp. &lat-LSTM v &28.1 $\pm$ 3.7 &40.0 $\pm$ 2.7 &50.0 $\pm$ 13.7 &50.6 $\pm$ 20.9 &42.0 $\pm$ 29.4 &41.2 $\pm$ 20.0 &56.0 $\pm$ 13.1 &64.9 $\pm$ 13.4 &63.6 $\pm$ 17.0 \\

comp. ind. &lat-LSTM e


&6.1 $\pm$ 4.3 &9.8 $\pm$ 6.4 &11.5 $\pm$ 7.6 &11.3 $\pm$ 7.6 &20.6 $\pm$ 0.8 &32.3 $\pm$ 7.9 &39.8 $\pm$ 15.1 &39.4 $\pm$ 20.1 &53.8 $\pm$ 25.2 
         \\   

       \rowcolor{gray!20} 

 comp. ind. &lat-LSTM e&40.8$\pm$ 28.0 &49.1$\pm$ 34.9 &52.9$\pm$ 38.0 &50.9$\pm$ 38.6 &51.3$\pm$ 34.1 &54.3$\pm$ 29.1 &59.6$\pm$ 26.1 &64.0$\pm$ 28.4 &69.6$\pm$ 34.4 &68.6$\pm$ 41.3

 \\
comp. ind. &lat-LSTM v &34.9 $\pm$ 4.7 &38.4 $\pm$ 2.3 &38.1 $\pm$ 6.2 &31.9 $\pm$ 4.1 &32.3 $\pm$ 3.7 &41.5 $\pm$ 8.0 &53.1 $\pm$ 11.8 &56.1 $\pm$ 1.7 &67.7 $\pm$ 13.8 
 \\

        comp. csp &conv-LSTM e& 100.0 $\pm$ 0.0 &0.0 $\pm$ 0.0 &0.0 $\pm$ 0.0 &0.0 $\pm$ 0.0 &\textbf{0.0 $\pm$ 0.0} &\textbf{0.0 $\pm$ 0.0} &\textbf{0.0 $\pm$ 0.0} &\textbf{0.0 $\pm$ 0.0} &\textbf{0.0 $\pm$ 0.0}  \\
        comp. csp &conv-LSTM v& 100.0 $\pm$ 0.0 &0.0 $\pm$ 0.0 &0.0 $\pm$ 0.0 &0.0 $\pm$ 0.0 &0.0 $\pm$ 0.0 &0.0 $\pm$ 0.0 &0.0 $\pm$ 0.0 &0.0 $\pm$ 0.0 &0.0 $\pm$ 0.0  \\

        \bottomrule
    \end{tabular}  
    \label{tab:fprpole}
\end{table*}








\begin{table*}[th]    
\caption{Brier score comparison before (white rows) and after (gray rows) calibration for the racing car with selected calibrators by hyperparameter-
tuning.}
    \centering
        \tiny 

    \begin{tabular}{llllllllllll}
        \toprule
       Predictor & Model & $k=20$ & $k=40$ & $k=60$ & $k=80$ & $k=100$ & $k=120$ & $k=140$ & $k=160$ & $k=180$ & $k=200$\\

        \midrule

mono. csp.& lat-LSTM&.0327$\pm$.0509  &.0977$\pm$.0715  &.1435$\pm$.0973  &.1796$\pm$.0950  &.2013$\pm$.1191  &.2192$\pm$.1264  &.2328$\pm$.1150  &.2409$\pm$.1161  &.2492$\pm$.1284  &.2689$\pm$.1307

\\
 \rowcolor{gray!20}
mono. csp.& lat-LSTM& .0294$\pm$.0446  &.0859$\pm$.0578  &.1167$\pm$.0698  &.1481$\pm$.0732  &.1524$\pm$.0767  &.1613$\pm$.0756  &.1741$\pm$.0693  &.1788$\pm$.0702  &.1858$\pm$.0681  &.1949$\pm$.0591

\\
mono. ind.& lat-LSTM& .0268$\pm$.0466  &.0924$\pm$.0978  &.1391$\pm$.1050  &.1669$\pm$.1086  &.1882$\pm$.1142  &.2066$\pm$.1246  &.1940$\pm$.1174  &.2173$\pm$.1273  &.2130$\pm$.1168  &.2146$\pm$.1181
 
\\
 \rowcolor{gray!20}

mono. ind.& lat-LSTM& \textbf{.0232$\pm$.0391 } &.0790$\pm$.0654  &.1105$\pm$.0755  &.1428$\pm$.0832  &.1516$\pm$.0850  &.1557$\pm$.0872  &\textbf{.1603$\pm$.0882}  &.1652$\pm$.0854  &\textbf{.1670$\pm$.0879}  &\textbf{.1714$\pm$.0875}

\\
   mono. csp.& CNN&.0315$\pm$.0496  &.0938$\pm$.0992  &.1276$\pm$.0974  &.1644$\pm$.1083  &.1771$\pm$.1083  &.1870$\pm$.1067  &.1945$\pm$.1019  &.2028$\pm$.1022  &.2137$\pm$.1082  &.2451$\pm$.1328

\\
 \rowcolor{gray!20}

mono. csp.& CNN& .0280$\pm$.0430  &.0790$\pm$.0748  &\textbf{.1066$\pm$.0799 } &\textbf{.1373$\pm$.0804 } &\textbf{.1431$\pm$.0843}  &\textbf{.1517$\pm$.0860}  &.1615$\pm$.0869  &\textbf{.1650$\pm$.0825}  &.1684$\pm$.0844  &.1743$\pm$.0804

\\
mono. ind.& CNN &.0359$\pm$.0545  &.0931$\pm$.0832  &.1361$\pm$.0944  &.1874$\pm$.1028  &.1924$\pm$.0988  &.1982$\pm$.1111  &.2067$\pm$.1038  &.2047$\pm$.1011  &.2106$\pm$.1114  &.2218$\pm$.1040

\\
 \rowcolor{gray!20}

mono. ind.& CNN& .0314$\pm$.0455  &\textbf{.0765$\pm$.0739 } &.1116$\pm$.0785  &.1488$\pm$.0769  &.1575$\pm$.0738  &.1607$\pm$.0838  &.1690$\pm$.0856  &.1753$\pm$.0749  &.1777$\pm$.0812  &.1823$\pm$.0804

\\
comp. csp. & lat-LSTM& .0341$\pm$.0134  &.1465$\pm$.0368  &.2303$\pm$.0509  &.2966$\pm$.0567  &.3139$\pm$.0636  &.3408$\pm$.0651  &.3626$\pm$.0701  &.3778$\pm$.0720  &.4099$\pm$.0782  &.4402$\pm$.0836

\\
 \rowcolor{gray!20}

comp. csp. & lat-LSTM& .0335$\pm$.0128  &.1446$\pm$.0403  &.2270$\pm$.0562  &.2797$\pm$.0689  &.3087$\pm$.0727  &.3231$\pm$.0808  &.3416$\pm$.0789  &.3682$\pm$.0900  &.3981$\pm$.0978  &.4269$\pm$.1092

\\
comp. ind. & lat-LSTM &.0368$\pm$.0158  &.1157$\pm$.0424  &.1681$\pm$.0576  &.2158$\pm$.0691  &.2539$\pm$.0782  &.2922$\pm$.0883  &.3267$\pm$.0981  &.3518$\pm$.1078  &.3752$\pm$.1081  &.4079$\pm$.1162

\\
 \rowcolor{gray!20}

comp. ind. & lat-LSTM& .0359$\pm$.0147  &.1145$\pm$.0431  &.1399$\pm$.0524  &.1671$\pm$.0593  &.2131$\pm$.0796  &.2495$\pm$.0848  &.2853$\pm$.0959  &.3185$\pm$.1069  &.3418$\pm$.1152  &.3693$\pm$.1228

\\
        \bottomrule
    \end{tabular}  
    \label{tab:briercar}
\end{table*}

\begin{table*}[th]
    \caption{Brier score comparison before (white rows) and after (gray rows) calibration for the cartpole with selected calibrators by hyperparameter-
tuning.}
    \centering
    \tiny 
    \begin{tabular}{llllllllllll}
        \toprule
        Predictor & Model & $k=3$ & $k=6$ & $k=9$ & $k=12$ & $k=15$ & $k=18$ & $k=21$ & $k=24$ & $k=27$ \\
        \midrule

    mono. csp. & lat-LSTM  
&.0021$\pm$.0037  &.0020$\pm$.0035  &.0015$\pm$.0022  &.0024$\pm$.0045  &.0039$\pm$.0069  &.0134$\pm$.0170  &.0693$\pm$.0541  &.1552$\pm$.0657  &.1978$\pm$.0373
\\
       \rowcolor{gray!20} 
        mono. csp. & lat-LSTM  &\textbf{.0021$\pm$.0038}  &\textbf{.0019$\pm$.0034 } &.0015$\pm$.0022  &\textbf{.0023$\pm$.0042 } &\textbf{.0037$\pm$.0063 } &\textbf{.0126$\pm$.0155 } &.0550$\pm$.0403  &.1211$\pm$.0624  &.1270$\pm$.0567

\\
 mono. ind. & lat-LSTM
&.0025$\pm$.0038  &.0019$\pm$.0040  &.0011$\pm$.0023  &.0024$\pm$.0044  &.0051$\pm$.0086  &.0135$\pm$.0166  &.0683$\pm$.0554  &.1547$\pm$.0673  &.2014$\pm$.0374

\\

    \rowcolor{gray!20} 
        mono. ind. & lat-LSTM  &.0025$\pm$.0036  &.0019$\pm$.0038  &\textbf{.0011$\pm$.0022}  &.0024$\pm$.0044  &.0048$\pm$.0077  &.0127$\pm$.0153  &\textbf{.0539$\pm$.0407}  &.1191$\pm$.0612  &.1285$\pm$.0588
\\
mono. csp. & CNN &.4841$\pm$.4809  &.4846$\pm$.4811  &.4850$\pm$.4799  &.4845$\pm$.4777 
&.4848$\pm$.4658  &.4833$\pm$.4331  &.4856$\pm$.3878  &.4860$\pm$.3409  &.4893$\pm$.2910

\\

        \rowcolor{gray!20} 
        mono. csp. & CNN  &.0029$\pm$.0053  &.0030$\pm$.0059  &.0045$\pm$.0088  &.0059$\pm$.0117  &.0182$\pm$.0276  &.0467$\pm$.0455  &.0783$\pm$.0579  &.1068$\pm$.0615  &\textbf{.1228$\pm$.0380}
\\
        mono. ind. & CNN  
&.4889$\pm$.4813  &.4874$\pm$.4817  &.4862$\pm$.4773  &.4855$\pm$.4782  &.4857$\pm$.4637  &.4858$\pm$.4328  &.4881$\pm$.3866  &.4878$\pm$.3404  &.4876$\pm$.2907
\\

       \rowcolor{gray!20} 
        mono. ind. & CNN  &.0066$\pm$.0088  &.0060$\pm$.0097  &.0079$\pm$.0149  &.0059$\pm$.0112  &.0157$\pm$.0213  &.0474$\pm$.0440  &.0794$\pm$.0579  &\textbf{.1039$\pm$.0532}  &.1269$\pm$.0446

\\
        comp. ind. & lat-LSTM 
&.0440$\pm$.0652  &.0570$\pm$.0751  &.0777$\pm$.0773  &.0820$\pm$.0606  &.0900$\pm$.0684  &.1099$\pm$.0794  &.1420$\pm$.0983  &.1533$\pm$.1063  &.2174$\pm$.1165  
\\
    \rowcolor{gray!20} 
        comp. ind. & lat-LSTM 
&.0327$\pm$.0436  &.0435$\pm$.0518  &.0638$\pm$.0588  &.0698$\pm$.0505  &.0763$\pm$.0544  &.0912$\pm$.0608  &.1060$\pm$.0599  &.1109$\pm$.0590  &.1428$\pm$.0580 

\\
        comp. csp. & lat-LSTM 

&.0464$\pm$.0678  &.0691$\pm$.0703  &.1020$\pm$.0866  &.1440$\pm$.1427  &.1562$\pm$.1649  &.1527$\pm$.1369  &.1857$\pm$.1502  &.2201$\pm$.1510  &.2257$\pm$.1677 

\\
    \rowcolor{gray!20} 
        comp. csp. & lat-LSTM 
&.0356$\pm$.0466  &.0577$\pm$.0519  &.0784$\pm$.0605  &.0877$\pm$.0678  &.0891$\pm$.0674  &.0976$\pm$.0558  &.1133$\pm$.0570  &.1297$\pm$.0528  &.1254$\pm$.0548

        \\
        \bottomrule
    \end{tabular}
    \label{tab:brierpole}
\end{table*}

The following figures and tables with results are provided: 
\begin{itemize}
    \item The performance of our evaluators and VAEs (Tab.~\ref{tab:eval-vae-performance})
    \item The results for every controller-specific label predictor for both case studies (Tab. 5, 6, 7 and 8)
        \item Brier score comparison between before calibration and after calibration for the  racing car and cartpole (Tab. 9 and 10)
\end{itemize}

\newpage

\begin{table}[th]	
\caption{Hyperparameters of prediction and training}
	\centering
	\begin{tabular}{llllll}
		\toprule
		\cmidrule(r){1-6}
		Name     & m  & k&N&M &Q\\
		\midrule
		Racing car  &15  &10-200 & 1000 &200 &10 \\
		Cart pole     & 5 &  3-27    & 100 & 100 & 10 \\
		\bottomrule
	\end{tabular}  
	\label{tab:hyperparams}
\end{table}

\begin{table}[h]
	\caption{Performance of evaluators and VAEs}
	\centering
	\begin{tabular}{llll}
		\toprule
		\multicolumn{2}{c}{\textbf{Evaluator}}                   \\
		\cmidrule(r){1-4}
		Case     & Type   & Acc (\%) &FPR(\%)\\
		\midrule
		\multirow{1}{*}{Racing car} &CNN   & 99.747  &0.535   \\
		 \cline{1-4} \\

  		\multirow{1}{*}{Cart pole}  
		&CNN     &  98.845  &  0.762    \\
  \toprule
		\multicolumn{2}{c}{\textbf{VAE}}                   \\
  		\cmidrule(r){1-4}

    Case     & Type& Recon loss    &Acc (\%)\\
		\midrule
		\multirow{2}{*}{Racing car} &w/ Safety loss   & 14.769  &96.743  \\
		&  w/o Safety loss     &14.770  &    96.822   \\ \cline{1-4} \\

  		\multirow{2}{*}{Cart pole} &w/ Safety loss   &  3.983  &  97.225 \\
		&w/o Safety loss     & 3.977 &   97.605    \\

		\bottomrule
	\end{tabular}
	\label{tab:eval-vae-performance}
\end{table}

 \begin{table*}
\centering
\begin{tabular}{|c|c|c|c|}
\hline
\textbf{Layer Name} & \textbf{Layer Type} & \textbf{Parameter Settings} & \textbf{Output Size} \\
\hline
\multirow{2}{*}{Conv1} & Conv2D & Channels: 1 $\rightarrow$ 6, Kernel: 3x3 & (6,62,62) \\
\cline{2-4}
 & MaxPool2D & Kernel: 2x2, Stride: 2 & (6,31,31) \\
\hline
\multirow{2}{*}{Conv2} & Conv2D & Channels: 6 $\rightarrow$ 16, Kernel: 5x5 & (16,27,27) \\
\cline{2-4}
 & MaxPool2D & Kernel: 2x2, Stride: 2 &  (16,13,13)\\
\hline
FC1 & Linear & Input Size: 2704, Output Size: 1024 & (1,1024) \\
\hline
FC2 & Linear & Input Size: 1024, Output Size: 256 & (1,256) \\
\hline
FC3 & Linear & Input Size: 259, Output Size: 2 & (1,2) \\
\hline
Softmax & Softmax & (1,2) & Class Probability Distribution \\
\hline
\end{tabular}

\caption{Architecture of our CNNs (evaluators, single-image predictors)}
\label{tab:architecture-cnn}
\end{table*}

 \begin{table*}
    \centering
    \begin{tabular}{|c|c|c|}
        \hline
        \textbf{Layer Name} & \textbf{Layer Type} & \textbf{Output Size} \\
        \hline
        Input & - & $(\text{in\_channels}, \text{height}, \text{width})$ \\
        \hline
        Convolution & Conv2D & $(4 \times \text{out\_channels}, \text{height}, \text{width})$ \\
        \hline
        Concatenation & Concatenate & $(4 \times \text{out\_channels} + \text{in\_channels}, \text{height}, \text{width})$ \\
        \hline
        Chunking & Chunk & 4 tensors of size $(\text{out\_channels}, \text{height}, \text{width})$ \\
        \hline
        Input Gate & Sigmoid & $(\text{out\_channels}, \text{height}, \text{width})$ \\
        \hline
        Forget Gate & Sigmoid & $(\text{out\_channels}, \text{height}, \text{width})$ \\
        \hline
        Current Cell Output & Element-wise operations & $(\text{out\_channels}, \text{height}, \text{width})$ \\
        \hline
        Output Gate & Sigmoid & $(\text{out\_channels}, \text{height}, \text{width})$ \\
        \hline
        Current Hidden State & Element-wise operations & $(\text{out\_channels}, \text{height}, \text{width})$ \\
        \hline
    \end{tabular}
    \caption{Architecture of a cell of our image forecaster (convolutional LSTM, in\_channels and out\_channels mean the size of input channel and output channel; height and width are the size of input images)}
    \label{tab:architecture-conv-lstm}
\end{table*}

\begin{table*}
    \centering
    \begin{tabular}{|c|c|c|c|}
        \hline
        \multirow{2}{*}{Layer} & \multirow{2}{*}{Input Shape} & \multirow{2}{*}{Output Shape} & \multirow{2}{*}{Other} \\
         &  &  & \\
        \hline
        LSTM & $(\text{seq\_len}, \text{bs}, \text{latents} + \text{actions})$ & $(\text{seq\_len}, \text{bs}, \text{hiddens})$ & hiddens=256 \\
        \hline
        Linear & $(\text{seq\_len} \times \text{bs}, \text{hiddens})$ & $(\text{seq\_len} \times \text{bs}, 2)$ & - \\
        \hline
        Softmax & $(\text{seq\_len} \times \text{bs}, 2)$ & $(\text{seq\_len} \times \text{bs}, 2)$ & - \\
        \hline
    \end{tabular}
    \caption{Architecture of our latent LSTM (monolithic predictor, seq\_len, bs, hidden, latents and actions stand for the length of the input sequence, batchsize, size of hidden layer, size of latent vector and action data)}
    \label{tab:architecture-latent-lstm}
\end{table*}

\begin{table*}
    \centering
    \begin{tabular}{|c|c|c|}
        \hline
        \textbf{Layer Name} & \textbf{Layer Type} & \textbf{Output Size} \\
        \hline
        Input & - & $(\text{seq\_len}, \text{bs}, \text{latents} + \text{actions})$ \\
        \hline
        LSTM & LSTM & $(\text{seq\_len}, \text{bs}, \text{hiddens})$ \\
        \hline
        Linear & Linear & $(\text{seq\_len}, \text{bs}, 64)$ \\
        \hline
        Output & - & $(\text{seq\_len}, \text{bs}, 64)$ \\
        \hline
    \end{tabular}
    \caption{Architecture of our latent forecaster (LSTM, seq\_len, bs, hidden, latents and actions stand for the length of the input sequence, batchsize, size of hidden layer, size of latent vector and action data)}
    \label{tab:architecture-comp-lat}
\end{table*}

\begin{table*}
    \centering
    \begin{tabular}{|c|c|c|}
        \hline
        \textbf{Component} & \textbf{Layer Type} & \textbf{Output Size} \\
        \hline
        Encoder & - & $(\text{img\_channels}, \text{height}, \text{width})$ \\
        \hline
         & Conv2D & $(32, \text{height}/2, \text{width}/2)$ \\
         & Conv2D & $(64, \text{height}/4, \text{width}/4)$ \\
         & Conv2D & $(128, \text{height}/8, \text{width}/8)$ \\
         & Conv2D & $(256, \text{height}/16, \text{width}/16)$ \\
         & Linear & $(\text{latent\_size})$ \\
         & Linear & $(\text{latent\_size})$ \\
        \hline
        Decoder & - & $(\text{img\_channels}, \text{height}, \text{width})$ \\
        \hline
         & Linear & $(1024)$ \\
         & ConvTranspose2D & $(128, \text{height}/8, \text{width}/8)$ \\
         & ConvTranspose2D & $(64, \text{height}/4, \text{width}/4)$ \\
         & ConvTranspose2D & $(32, \text{height}/2, \text{width}/2)$ \\
         & ConvTranspose2D & $(\text{img\_channels}, \text{height}, \text{width})$ \\
        \hline
    \end{tabular}
    \caption{Architecture of VAEs (img\_channels, latent\_size, height and width are the channel of images, size of latent vector, the height and width of input images)}
    \label{tab:architecture-vae}
\end{table*}

\end{document}